\documentclass[lettersize,journal]{IEEEtran}
\usepackage{amsmath,amssymb,amsfonts}
\usepackage[ruled,linesnumbered]{algorithm2e}
\usepackage{array}
\usepackage[caption=false,font=normalsize,labelfont=sf,textfont=sf]{subfig}
\usepackage{textcomp}
\usepackage{stfloats}
\usepackage{url}
\usepackage{verbatim}
\usepackage{graphicx}
\usepackage{booktabs}
\usepackage{multirow}
\usepackage[numbers,sort&compress]{natbib}

\usepackage[colorlinks,linkcolor=blue,anchorcolor=black,citecolor=blue]{hyperref}
\begin{document}

\title{A Data-driven Crowd Simulation Framework Integrating Physics-informed Machine Learning with Navigation Potential Fields}

\author{
        Runkang~Guo,\IEEEmembership{}
        Bin~Chen,\IEEEmembership{}
        Qi~Zhang,\IEEEmembership{}
        Yong~Zhao,\IEEEmembership{}
        Xiao~Wang,\IEEEmembership{~Senior Member,~IEEE,}
        and Zhengqiu~Zhu* \IEEEmembership{}
\thanks{Manuscript received August XX, 2024; revised August XX, 2024. This study is supported by the National Natural Science Foundation of China 62173337 and Youth Independent Innovation Foundation of NUDT (ZK-2023-21). (\textit{Runkang Guo and Bin Chen contributed equally to this work}.) 
(\textit{Corresponding author: Zhengqiu Zhu}).}
\thanks{Runkang Guo, Bin Chen, Qi Zhang, Yong Zhao, and Zhengqiu Zhu are with the College of Systems Engineering, National University of Defense Technology, Changsha 410073, Hunan Province, China. 
(e-mail:~guorunkangnudt@nudt.edu.cn;~nudtcb9372@gmail.com;~zhangqiy123@nudt.edu.cn; \\
~zhaoyong15@nudt.edu.cn;~zhuzhengqiu12@nudt.edu.cn).}
\thanks{Xiao Wang is with the School of artificial intelligence, Anhui University of China, Hefei 230039, Anhui Province, China. (e-mail:~xiao.wang@ahu.edu.cn).}
}



\maketitle
\begin{abstract}
Crowd simulation plays a crucial role in the prevention and management of urban public safety events. However, the complexity and heterogeneity of human behaviors pose significant challenges to achieving realistic and accurate simulations. Traditional rule-based physical models are limited by their reliance on
singular physical formulas and parameters, making it difficult
to effectively tackle the intricate tasks associated with crowd
simulation. Recent research has introduced deep learning methods to tackle these issues, but most current approaches focus primarily on generating pedestrian trajectories, often lacking interpretability and failing to provide real-time dynamic simulations. 
To address the aforementioned issues, we propose a novel data-driven crowd simulation framework that integrates Physics-informed Machine Learning (PIML) with navigation potential fields. Our approach leverages the strengths of both physical models and PIML. Specifically, we design an innovative Physics-informed Spatio-temporal Graph Convolutional Network (PI-STGCN) as a data-driven module to predict pedestrian movement trends based on crowd spatio-temporal data. Additionally, we construct a physical model of navigation potential fields based on flow field theory to guide pedestrian movements, thereby reinforcing physical constraints during the simulation. In our framework, navigation potential fields are dynamically computed and updated based on the movement trends predicted by the PI-STGCN, while the updated crowd dynamics, guided by these fields, subsequently feed back into the PI-STGCN. Comparative experiments on two publicly available large-scale real-world datasets across five scenes demonstrate that our proposed framework outperforms existing rule-based methods in accuracy and fidelity. The similarity between simulated and actual pedestrian trajectories increases by 10.8\%, while the average error is reduced by 4\%. Moreover, our framework exhibits greater adaptability and better interpretability compared to methods that rely solely on deep learning for trajectory generation.
\end{abstract}


\section{Introduction}
\IEEEPARstart{C}{rowd} simulation is a technique that simulates the dynamics of virtual individual movements~\cite{1,yang2020pedestrian}, extensively utilized in fields such as computer graphics~\cite{2,3}, computer games~\cite{4,5}, urban planning~\cite{6,duan2022modeling}, and emergency management~\cite{9,10}. With the progression of global urbanization, urban populations are increasing, resulting in a rise in public safety incidents. Consequently, research on how to simulate real-world individuals and enhance the authenticity of crowd simulation holds paramount importance for effective urban planning and emergency evacuation.

Existing crowd simulation methods can generally be divided into rule-based methods and data-driven methods~\cite{63}. Rule-based methods typically use domain expertise or specific rules to construct crowd simulation models. Widely utilized in a variety of crowd simulation tasks, many of these rule-based methods demonstrate exceptional results in specific aspects. 
For instance, in collision avoidance, the Optimal Reciprocal Collision Avoidance (ORCA)~\cite{14,15} algorithm demonstrated outstanding performance. In collective behavior, improved algorithms based on classical social force models~\cite{17} were successfully applied in high-density crowd simulation and evacuation scenarios, accurately replicating typical crowd evacuation phenomena, such as competition and cooperation~\cite{19,20}. To study the psychological and behavioral characteristics of crowds, agent-based methods are widely used. These methods~\cite{21} often assume that each agent is an independent entity equipped with perception, decision-making, and action capabilities, thereby enabling the simulation of complex human behavior at both local and global levels. However, developing behavior rules for agents to consistently generate realistic movements is challenging. In addition, global path planning~\cite{22} is costly for each agent, especially in environments with frequent pedestrian interactions. Conversely, the potential field model~\cite{24,25} enables rapid global path planning by representing crowds as fields or flows. Especially in scenarios with large-scale crowds, the computational complexity of potential field is low, making it highly suitable for real-time crowd simulation.

Human behavior is inherently complex, and the psychological interactions within crowds are difficult to quantify using simple rules. Representing the complex dynamics of crowds solely through homogeneous rules or physical calculations is confronted with many issues. Additionally, rule-based methods heavily rely on empirical knowledge and require the adjustment of numerous parameters to achieve accurate simulation~\cite{13}. With advancements in sensing technologies, obtaining crowd trajectory data has become easier, paving the way for the development of data-driven crowd simulation methods. Early data-driven approaches primarily focused on constructing crowd databases~\cite{26} and performing data analysis based on statistical learning~\cite{27}, which made adaptation to environments with complex interactions difficult. In recent years, advancements in data science and deep learning have significantly heightened the research focus on crowd simulation based on deep learning~\cite{28}. Researchers employ various deep neural networks to extract spatio-temporal features of pedestrian trajectories and construct simulations by predicting pedestrian behaviors, primarily utilizing methods such as Long Short-Term Memory networks (LSTM)~\cite{31}, Generative Adversarial Networks (GAN)~\cite{30}, Graph Convolutional Networks (GCN)~\cite{32}, and Transformer models~\cite{29}. 

Although these methods can generate relatively accurate simulated trajectories, their black-box nature results in a lack of interpretability. Additionally, as the extrapolation prediction horizon extends, neural network models exhibit a super-linear increase in simulation errors~\cite{33}. Consequently, recent studies integrated deep learning with rule-based methods to implement crowd simulations. For instance, Li \textit{et al}. combined deep learning with the ORCA algorithm to enhance the fidelity of crowd simulation~\cite{34}, while Zhang \textit{et al}. employed deep learning to construct a neural network resembling a social force model for crowd simulation~\cite{35}. However, these hybrid methods still heavily depend on the design of neural networks to directly predict physical model parameters, which poses challenges in simulating crowd behaviors beyond the distribution of the training data.

To address these challenges, we propose a data-driven crowd simulation framework integrating Physics-informed Machine Learning (PIML)~\cite{36} with navigation potential fields. This framework integrates deep learning with physical models and employs the agent-based method to simulate heterogeneous crowds. Our objective is to improve the accuracy and fidelity of crowd simulation by creating a closed data-driven simulation loop, thereby providing robust computational support for crowd management and intervention. In the proposed framework, we design a Physics-informed Spatio-temporal Graph Convolutional Network (PI-STGCN) to predict pedestrian movement trends in real-time under the constraints of motion physics, which addresses the challenges of low heterogeneity in rule-based models and poor interpretability of purely machine learning methods. Moreover, to overcome the limitations of existing data-driven simulations regarding generalization and their inadequacy for long-term crowd simulation, we propose to employ data-driven navigation potential fields to guide pedestrian movements. This approach transforms movement trends learned from trajectory data into navigation fields that effectively direct pedestrian motion. Consequently, our crowd simulation framework adapt well to complex obstacle scenarios and pedestrian interactions.

In summary, the novel contributions of this paper are as follows:

\begin{enumerate}
\item A novel data-driven crowd simulation framework that integrates PIML with navigation potential fields is proposed. This framework combines deep learning methods with physical models, creating a closed-loop computation that fuses data and physics. Extensive experiments on two large-scale public datasets demonstrate the effectiveness of our proposed framework. compared to those generated by rule-based models, the crowd movements simulated based on the proposed framework are more accurate and reasonable. Moreover, our framework provides greater interpretability than existing deep learning methods, making it more suitable for long-term, real-time crowd simulation tasks.
\item A pedestrian movement trend prediction model based on PI-STGCN is developed, realized through innovative designs. To address the lack of physical constraints in traditional deep learning methods, based on the Spatio-temporal Graph Convolutional Network (STGCN) to capture spatio-temporal dependencies in pedestrian trajectories, a Physics-informed Neural Network (PINN)~\cite{37} is designed as a surrogate model for the physical relationship between pedestrian displacement and velocity. The PINN is connected to the STGCN by receiving it's displacement encoding output, thereby forming the PI-STGCN. In the PI-STGCN, a physical loss function and a velocity loss function are constructed to embed the physical equations of displacement and velocity into the training of the neural network. By leveraging PI-STGCN, the movement trends of pedestrians can be predicted accurately in real-time during crowd simulation. 
\item A data-driven algorithm for modeling navigation potential fields is designed.
To address the poor generalization of existing deep learning methods for directly predicting trajectories in long-term simulations, we introduce navigation potential fields based on flow field theory to represent the short-term movement trends of pedestrians. Furthermore, through a novel numerical computation approach, the navigation potential fields incorporate guidance information to help agents avoid obstacles and interact with other pedestrians, enabling efficient and real-time path planning for heterogeneous pedestrians during simulation. Analysis of simulated trajectories derived from extensive experiments reveals that movements based on the navigation potential fields exhibit advanced adaptability to complex scenarios and improved interpretability.
\end{enumerate}

The remainder of this paper is structured as follows: Section II reviews related work. Section III outlines the proposed crowd simulation framework, the architecture of the PI-STGCN, and navigation potential fields. Section IV details the construction method of the navigation potential fields. Section V presents the experiment process and discusses the results. Finally, Section VI provides our conclusions and suggestions for future work.

\section{Ralated Work}
\subsection{Applying Rule-Based Methods in Crowd Simulation}
Rule-based crowd simulation methods can be categorized into macroscopic models and microscopic models~\cite{1}. At the macroscopic level, crowd simulation algorithms emphasize group path planning or global control, employing methods such as the continuum model~\cite{58}, the aggregate dynamics model~\cite{59}, and potential field~\cite{24,25} to guide group movement. Conversely, at the microscopic level, most crowd simulation models are agent-based, focusing on individual agent characteristics and interactions between agents. These microscopic models involve modeling different behaviors of agents based on attributes such as velocity~\cite{60}, visual properties~\cite{61}, or psychological properties~\cite{62}, establishing rules for each agent to implement crowd simulation. Research in this field primarily focuses on local interaction behaviors among agents, with collision avoidance being a primary concern. The most widely used velocity-based collision avoidance models were developed by a research group from the University of North Carolina. These includes Velocity Obstacles (VO)~\cite{38}, Reciprocal Velocity Obstacles (RVO)~\cite{39}, Optimal Reciprocal Collision Avoidance (ORCA)~\cite{14}, and Hybrid Reciprocal Velocity Obstacles (HRVO) models~\cite{40}. In 2000, Helbing introduced the renowned social force model, which interprets the motion of each agent as a result of attraction towards goals, avoidance forces between agents, and repulsive interactions between agents and obstacles. In addition to avoiding collisions, many other types of local behaviors are considered, such as grouping and following. For instance, Qiu Modeled group structures in crowd simulation and analyzed the phenomenon of grouping within crowds~\cite{41}.

Rule-based crowd simulation methods abstract crowd movement into mathematical equations or deterministic system, demonstrating excellent scalability and stability under certain conditions. However, rule-based homogeneous behavior models rely on expert knowledge and rigorous hypothesis, making it difficult to capture the diverse and complex interactions of crowd movements in real situations. Therefore, rule-based models for crowd simulation are limited in fidelity and accuracy, while data-driven approaches can compensate for these limitations~\cite{wang2019social}.

\subsection{Applying Potential Field Models in Crowd Simulation}
In scenarios where individuals have global objectives, simulating real-world crowds solely based on collision avoidance is insufficient. Therefore, many crowd models integrate collision avoidance with global navigation, employing potential field methods to guide crowd movement. Chenney was the first to introduce a flow tiles model~\cite{43}, demonstrating that ideal crowd properties can be represented using divergence-free flows. In addition, motion flow fields extracted from crowd video footage were utilized to guide one or more agents toward desired goals~\cite{25}. In the context of crowd evacuation, Sun developed a density navigation field model based on potential field to investigate ways to enhance evacuation efficiency~\cite{42}.
Moreover, potential field models offer users the ability to dynamically create and control crowd simulations at runtime. The interactive navigation field~\cite{45} methods guide the motion of one or more agents by integrating user-defined guiding fields. 
Most of these methods provide control over global paths by allowing users to draw curves for agents to follow or by converting them into directional cues for the flow field.
For instance, Bönsch proposed a sketch-based VR method that provides options to redirect a flow by sketching barriers or guiding entities based on a sketched network of connected sections~\cite{46}. 
Additionally, others~\cite{48,64} also proposed sketch-based control of other simulation parameters, such as walking speed, the smoothness of paths, and the geometry of the environment.
Some commercial animation systems allow users to draw simple directional flow fields on the scene to guide the movement of individuals~\cite{49}.

Potential fields applied in crowd simulation exhibit high scalability and rapid global path planning, effectively adapting to a variety of complex crowd behaviors. However, most current research remains dependent on rule-based and subjectively defined potential fields, with few studies exploring data-driven methods to automatically update potential fields in simulation.

\vspace{-2pt}

\subsection{Applying Deep Learning Methods in Crowd Simulation}
The rapid advancement of artificial intelligence technology has made deep learning an essential tool for applications involving crowd simulation. Research on crowd trajectory prediction using deep learning leverages various neural network architectures to extract features from crowd trajectories. Some of these studies focused on the randomness of crowd behaviors by employing GAN to generate diverse crowd trajectories~\cite{30}. However, these approaches often face challenges, including unstable training processes, slow convergence, and limited interpretability. Other Methods, based on GCN, modeled pedestrians as nodes and represented the relationships between nodes as a graph network structure~\cite{50,51}. For instance, Social Spatio-temporal Graph Convolutional Neural Network (Social-STGCNN)~\cite{32} is a representative work for pedestrian trajectory prediction. This network embeds the social interaction relationships among pedestrians within an adjacency matrix and processes the matrix through kernel functions to capture both spatial and temporal information.
Most trajectory prediction methods based on deep learning primarily focus on generating the spatial position information of pedestrians, often neglecting the underlying physical mechanisms of pedestrian movement, which leads to poor interpretability. In response to this challenge, recent research developed deep learning methods that incorporate physical models~\cite{52,du2024social}. Given the widespread application of PINN in PIML, a network that surrogates social force models was developed to predict pedestrian movements~\cite{35}. In addition, Yu \textit{et al.} employed continuum models combined with neural networks to control crowd behaviors~\cite{53}, further demonstrating the efficacy of integrating traditional methods with deep learning for crowd simulation. Despite these advancements, current methods remain reliant on neural network designs, limiting their ability to adapt to simulations that fall outside the distribution of the training data.

Based on existing spatio-temporal graph neural network models, this paper employs the PIML framework to predict pedestrian movement trends and integrates it with a physical model of navigation potential fields, to implement data-driven crowd simulations.
\section{The Data-driven Crowd Simulation Framework}

\begin{figure*}[htbp]
	\centering
	\includegraphics[width=0.75\linewidth]{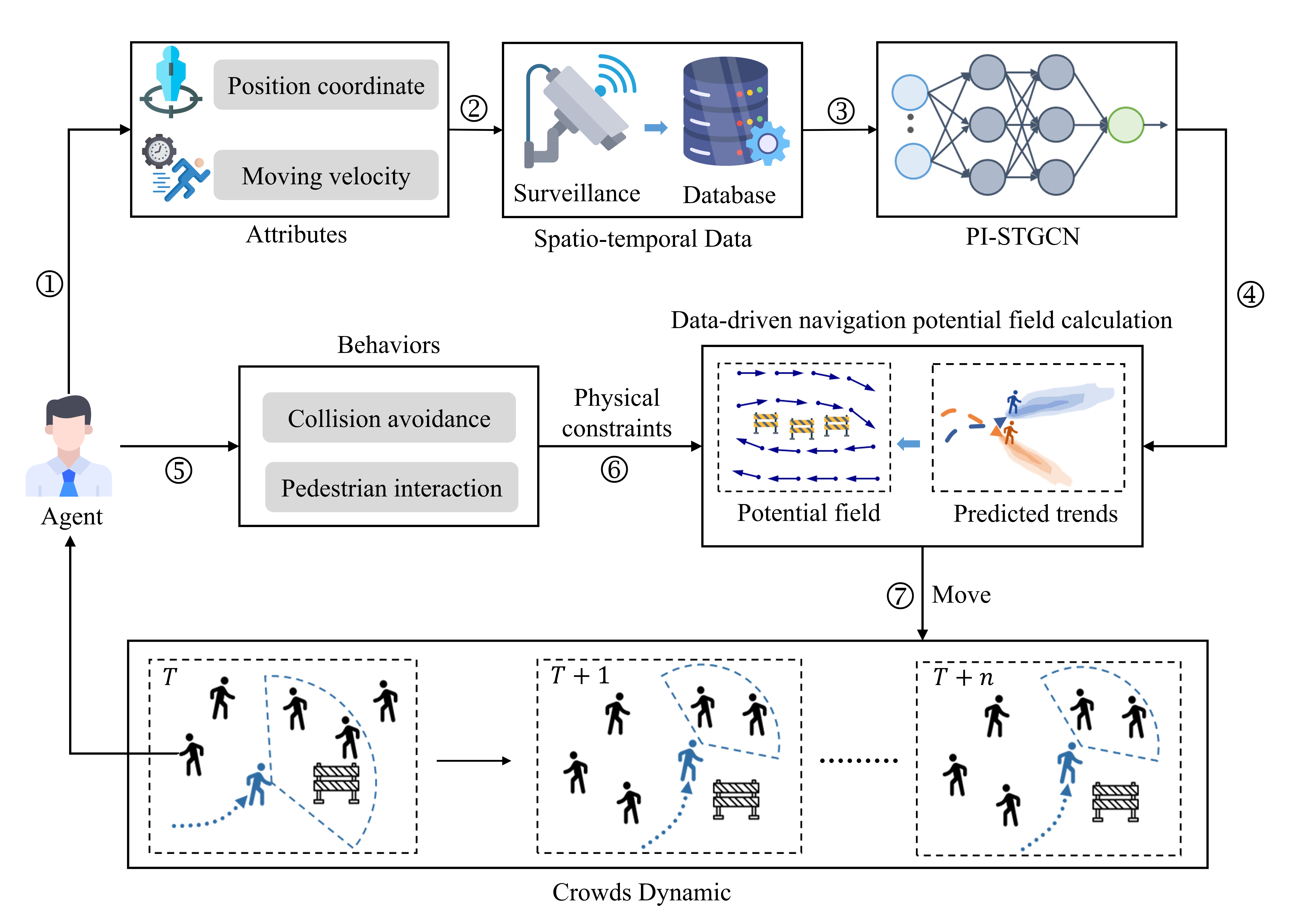}
        \centering
	\caption{A multi-agent crowd simulation framework based on Physics-informed Machine Learning and navigation potential fields.}
	\label{fig:1}
\end{figure*}

\subsection{Overall Framework}
As shown in Figure \ref{fig:1}, our crowd simulation framework adopts a hybrid architecture that combines physical models with deep learning models. In the framework, PIML functions as the data-driven module to predict pedestrian movement trends, while the potential fields serve as the underlying physical model that supports large-scale computation of path planning. The crowd simulation based on this framework consists of three stages: 
(i) Prediction of movement trends based on PI-STGCN. Crowd dynamics in simulation are continuously sensed by an intelligent surveillance system and stored in a database, creating dynamically updated spatio-temporal data of pedestrian trajectories. Subsequently, the PI-STGCN leverages the positions and velocities of pedestrians, along with physical dynamics from the spatio-temporal information, to train the neural network and predict the movement trends of agents in real time. 
(ii) Construction of data-driven navigation potential fields. The position distribution function predicted by PI-STGCN is sampled to generate movement trend lines. Taking into account agents' collision avoidance and pedestrian interaction behaviors, an algorithm is required to further convert these trend lines into numerical potential fields.
(iii) Execution of agents' movement. Under the influence of navigation potential fields, agents plan their paths in real time based on the distribution of potential values in the environment, and then execute movements according to the planned paths to update their positions. After all agents complete their actions within a simulation time step, the crowd dynamics are updated accordingly.
In the following sections, a detailed description of the specific models and computational processes will be explained.

\subsection{Attributes and Behaviors of Agents}
In this paper, we leverage an agent-based modeling method to depict heterogeneous crowds. In order to enhance computational efficiency, the simulation employs a semi-continuous computing method. This involves each individual moving a specific distance in their direction of motion at each discrete time step. 
In our framework, the environment is discretized into a two-dimensional grid plane, and then a corresponding two-dimensional potential field is established. Each pedestrian is represented by an agent that guided towards the goal by the potential field. As an agent moves, it dynamically adjusts velocity and direction based on the environment to avoid collisions with other pedestrians and obstacles.
The attributes of agent $i$ at time $t$ are:
\begin{enumerate}
\item Position coordinates: 
Assuming that the environment is represented by a Cartesian coordinate system, the position coordinates of pedestrian $i$ at time $t$ are represented as $S_{t}^{i}=(x_{t}^{i},y_{t}^{i})$.
\item Velocity vector: 
The velocity of the pedestrian is represented by $\vec{V}_{t}^{i}=\left(v_{x, t}^{i}, v_{y, t}^{i}\right)$ in the Cartesian coordinate system. The velocity angle $v_{\varphi}=\arctan \left(\frac{v_{x, t}^{i}}{v_{y, t}^{i}}\right)$ indicates the current direction of movement of the pedestrian. In a discretized grid environment, the velocity angle can take 8 possible values, each corresponding to one of the 8 neighboring grids towards which the pedestrian is moving.
\end{enumerate}

Each agent exhibits autonomous behaviors, which are divided into three stages: sensing, planning, and execution.
\subsubsection{Sensing}
It is assumed that pedestrians are capable of visually sense the current environment and dynamics of other pedestrians. After the position is updated, pedestrians sense obstacles and other pedestrians in the environment based on their current position and predetermined goals, providing decision-making information for planning their next move. In a discretized two-dimensional plane, the environmental properties are represented by matrix $\mathbf{E}$, where the value $e_{ij}$ at grid $(i, j)$ represents the type of passage. In this case, only two types are considered: 0 (obstacle) and 1 (passable). The crowd dynamics, denoted as $\mathbf{C}$, encapsulates the attribute information of all agents. At a given time $t$, the crowd dynamics is represented as $C_{t}=\big(S_{t}, \vec{V}_{t}\big)$. And we introduce a navigation potential field $\mathbf{F}$ that is mapped to the two-dimensional grid plane to depict the agents' movement tendencies. The detailed construction method with its process is described in Section IV. Therefore, for all agents in the environment at time $t$, they are able to sense environmental attributes, crowd behaviors, and movement trends within a certain range of their position, generating spatio-temporal information $I=\{\mathbf{E},\mathbf{C},\mathbf{F}\}$.

\subsubsection{Planning}
During the planning phase, the primary task is to calculate pedestrian movement routes. Traditional approaches that use shortest path global planning do not adequately reflect the heterogeneity and realistic tendencies of pedestrian behaviors. In contrast, this paper employs field methods to guide agents in selecting routes, representing a data-driven approach to local planning. Agents leverage real-time environmental information and crowd dynamics to update the navigation fields via predicting movement trends. By reinforcing physical constraints for obstacle avoidance and pedestrian interactions, realistic and feasible movement paths can be rapidly planned.

\subsubsection{Execution}
The agent updates its position after executing actions based on the calculated velocity and direction from the path planning. As the position is updated, the crowd dynamics change accordingly. In this paper, we assume that the passability of the environment remains constant. 
The update frequency of the navigation potential fields is tied to the data-driven period. If the movement trends are updated with each position update, the navigation potential fields will be updated in real time. However, considering the practical aspects of crowd movements and simulation efficiency, we assume that pedestrians' movement trends remain stable within a certain time interval. Therefore, after an agent completes an action, it enters a new phase of sensing, forming a closed loop of the data-driven simulation.

\subsection{Predicting Pedestrian Movement Trends Based on PIML}
In this section, the details of the PI-STGCN for predicting pedestrian movement trends are presented. Table \ref{tab:I} shows the main concepts and notations used in the PI-STGCN.

\begin{table}[!htb]
\centering
\caption{Main notations and concepts}
\label{tab:I}
\resizebox{\linewidth}{!}{
\begin{tabular}{cp{5.5cm} l}
\hline
\hline
\textbf{Notations}                & \textbf{Explanations} \\
\hline
$N()$               & A bivariate Gaussian distribution function \\ 
${S}$              & Agent position coordinates \\   
${V}$              & Agent velocity vector \\
${G_{t}^{S}}$              & Pedestrian displacement spatio-temporal graph \\
${G_{t}^{V}}$              & Pedestrian velocity spatio-temporal graph \\
${\mathbf{A}}$              & The adjacency matrix corresponding to the spatio-temporal graph \\
${\mathbf{D}}$              & The degree matrix corresponding to the adjacency matrix \\
${\hat{\mu}}$              & Mean of Gaussian distribution for predicting position \\
${\hat{\sigma}}$              & Covariance of Gaussian distribution for predicting position \\
${\hat{\rho}}$              & Correlation coefficient of Gaussian distribution for predicting position \\
${l_{p}}$              & Position loss function \\
${l_{f}}$              & Physical loss function \\
${l_{v}}$              & Velocity loss function \\
${\omega_{p}}$              & A hyperparameter in the neural network that represents the weight of the position loss \\
${\omega_{f}}$              & A hyperparameter in the neural network that represents the weight of the physical loss. \\
${\omega_{v}}$              & A hyperparameter in the neural network that represents the weight of the velocity loss \\
\hline
\hline
\end{tabular}}
\end{table}

\begin{figure*}[htbp]
	\centering
	\includegraphics[width=1\linewidth]{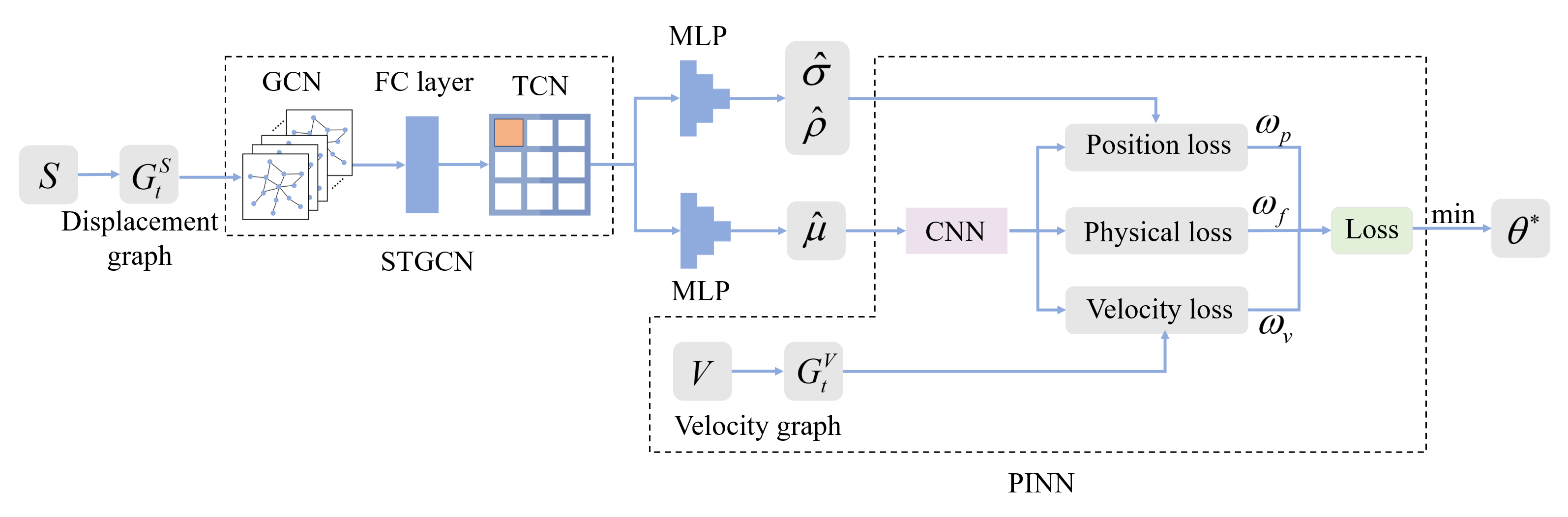}
        \centering
	\caption{The structure of the Physics-informed Spatio-temporal Graph Convolutional Network.}
	\label{fig:2}
\end{figure*}

The network structure of the PI-STGCN is illustrated in Figure \ref{fig:2}. The neural network receives as input the observed positions and velocities time series of agents. Through spatio-temporal graph convolution operations and the learning of physical information, it outputs the parameters of a bivariate Gaussian distribution function representing future movement trends.

The position of a pedestrian is represented by a bivariate Gaussian distribution function~\cite{31} in $x$ and $y$ directions. $\mu$ denotes the mean of the $x$ and $y$ coordinates, $\sigma$ represents the corresponding variance, and $\rho$ indicates the correlation coefficient between $x$ and $y$. Therefore, the distribution of the position can be expressed as:
\begin{equation}
S \sim N(\mu, \sigma, \rho)
\label{eq:gauss}
\end{equation}
A displacement graph ${G_{t}^{S}}$ is constructed based on the positions of all agents in the crowd at time $t$. Each vertex in the graph corresponds to an agent, with the vertex feature representing the pedestrian's displacement within a time step of $\Delta t$, denoted as $S_{t}-S_{\Delta t}$.
The connectivity between vertices is represented by an adjacency matrix ${\mathbf{A}}$. If the distance between pedestrian $i$ and pedestrian $j$ is not equal to 0, the corresponding element $a_{ij}$ is set to $a_{i j}=\frac{1}{\left\|S_{t}^{i}-S_{t}^{j}\right\|}$; Otherwise,  the two pedestrians are considered as the same agent, i.e., $a_{i j}=0$. Similarly, replacing the displacement features of the vertices in the graph with the pedestrian's velocity $V$, the velocity graph ${G_{t}^{V}}$ is constructed.

Graph ${G_{t}^{S}}$ is input of the spatio-temporal graph neural network. Let $\mathbf{H}^{l}$ be the displacement feature tensor at the $l$ layer. After one graph convolution, $\mathbf{H}^{l+1}$ is computed. The mathematical representation of the propagation process is:
\begin{equation}
\mathbf{H}^{l+1}=\text{ReLU}\left(\tilde{\mathbf{D}}^{-\frac{1}{2}} \tilde{\mathbf{A}} \tilde{\mathbf{D}}^{-\frac{1}{2}} \mathbf{H}^{(l)} \mathbf{W}^{(\mathbf{l})}\right)
\label{eq:gcn}
\end{equation}
Here, $\tilde{\mathbf{A}}=\mathbf{I}_{\mathbf{N}}+\mathbf{A}$ represents the normalized adjacency matrix, $\tilde{\mathbf{D}}_{i j}=\sum_{j} \tilde{\mathbf{A}}_{i j}$ is the normalized degree matrix, $\mathbf{W}^{(\mathbf{l})}$ is the graph convolution weight parameter matrix. The embedding vector obtained from the GCN output is passed through a fully connected network to compute the encoding vector $h^{l}$. $h^{l}$ encodes the hidden spatial dependencies between pedestrians. To uncover the hidden temporal dependencies within the data, a Temporal Convolutional Network (TCN) is leveraged to perform convolution operations on the temporal dimension of the sequence. Simultaneously, residual layers have been incorporated to improve the generalization capacity of neural network training. Following a single TCN computation, the output feature is $h^{l+1}=\text{ReLU}\left(h^{l}+\mathbf{W}^{(\mathbf{c})} h^{l}\right)$. $\text{ReLU()}$ represents the activation function, and $\mathbf{W}^{(\mathbf{c})}$ stands for the weight vector of the convolutional network.

In spatio-temporal GCN, the number of layers for GCN and TCN is determined based on the specific context. After performing multiple layers of GCN and TCN propagation calculations, the spatio-temporal encoding capturing the hidden spatio-temporal dependencies of pedestrian movements is obtained. We utilize a Multi-Layer Perceptron (MLP) for decoding, converting the encoded vector into parameters of predicted Gaussian distribution, and finally obtain initial parameters $\hat{\mu}$, $\hat{\sigma}$, and $\hat{\rho}$. Based on the Gaussian distribution function, the negative log-likelihood function~\cite{32} under the predicted parameter distribution is calculated as the loss function for predicting the position distribution, denoted as $l_{p}$. As shown in Equation \eqref{eq:lp}, $T_{p}$ represents the time length of the predicted position sequence.
\begin{equation}
l_{p}=-\sum_{t=1}^{T_{p}} \log \left(\mathbb{P}\left(S_{t} \mid \hat{\sigma}, \hat{\mu}, \hat{\rho}\right)\right)
\label{eq:lp}
\end{equation}

\begin{algorithm}[htbp]
\label{al:pi-stgcn}
\caption{The procedure of the PI-STGCN model.} 
\hangindent=4.5em 
        \hangafter=1
\KwIn{
The displacement graph ${G_{t}^{S}}$, \\ \hspace{3.1em} The velocity graph ${G_{t}^{V}}$, \\ \hspace{3.5em}and the training epochs $k$, $\omega_{p}$, $\omega_{f}$, $\omega_{v}$.}
\KwOut{Predicted position $\hat{\mu}$.}

    Initialize hidden state $\mathbf{H}^{\mathbf{0}}$, and parameters $\mathit{\theta}$;

    \For{$\textit{epoch}=1$ \textit{to} k}
    {
        \For{\textit{each layer in } STGCN}
        {
             $\mathbf{H}^{\mathbf{l + 1}} \leftarrow \text{ReLU}\left(\tilde{\mathbf{D}}^{-\frac{1}{2}} \tilde{\mathbf{A}} \tilde{\mathbf{D}}^{-\frac{1}{2}} \mathbf{H}^{\mathbf{l}} \mathbf{W}^{(\mathbf{l})}\right)$

             $h^{l} \leftarrow \mathbf{H}^{\mathbf{l}}$
             
             $\mathbf{H}^{\mathbf{l + 1}} \leftarrow \text{ReLU}\left(h^{l}+\mathbf{W}^{(\mathbf{c})}h^{l}\right)$
        }
        $\left(\hat{\mu}_{t}, \hat{\sigma}_{t}, \hat{\rho}_{t}\right) \leftarrow \text{MLP}(\mathbf{H}^{k})$
        
        $\hat{V} \leftarrow \text{PINN}(t,\hat{\mu})$
        
        Compute $l_{p}$, $l_{f}$, and $l_{v}$ by  Eq. \eqref{eq:lp}\eqref{eq:lf}\eqref{eq:lv};
        
        Compute the gradient of model parameters $\theta$; 
        
        Update $\theta$ according to loss with SGD optimizer;
    }
    
    \Return $\hat{\mu}$
\end{algorithm}
To introduce the physical information in the data-driven process, a PINN is constructed that embeds the displacement-velocity dynamics equations of pedestrian motion into the parameter updates of the neural network. Specifically, a multi-layer convolutional neural network is utilized as a surrogate model to represent the mapping relationship between pedestrian displacement and velocity. Then, after being mapped by the neural network, the parameter $\hat{\mu}$ yields the predicted velocity $V$. As the actual movement of pedestrians is a complex and nonlinear physical process, acceleration during movement varies irregularly over time. In order to incorporate the physical constraints of the motion into the PINN, under the assumption of discrete movement model, pedestrians are approximated as uniform acceleration within a small time interval~\cite{17}.
Based on the condition, two types of loss functions are designed to quantify the effectiveness of the neural network surrogate model. Firstly, an unsupervised loss function $l_{f}$ is defined based on the kinematic equations of uniform acceleration. Secondly, the observed velocity information is constructed a pedestrian velocity spatio-temporal graph to compute the supervised loss function $l_{v}$.

\begin{equation}
l_{f}=\sum_{t=1}^{T_{p}}\left(\left(\frac{1}{2} \times\left(\frac{\partial \hat{S}_{t}}{\partial t}+\frac{\partial \hat{S}_{t-1}}{\partial t}\right) \Delta t-\left(S_{t}-S_{t-1}\right)\right)^{2}\right)
\label{eq:lf}
\end{equation}

\begin{equation}
l_{v}=\sum_{t=1}^{T_{p}}\left(\left(\hat{V}_{t}-V_{t}\right)^{2}\right)
\label{eq:lv}
\end{equation}

Algorithm \ref{al:pi-stgcn} illustrates the training process of the PI-STGCN. To balance the varying impact of loss functions on the gradient updates of the neural network during the training phase, $\omega_{p}$, $\omega_{f}$, and $\omega_{v}$ are introduced as hyperparameters. The training objective of PI-STGCN is
\begin{equation}
\theta^{*}=\underset{\theta}{\arg \min }\left(\omega_{p} l_{p}+\omega_{f} l_{f}+\omega_{v} l_{v}\right)
\label{eq:train_obj}
\end{equation}

\section{Navigation Potential Fields}
This section describes the computational methods for constructing navigation potential fields based on the prediction of movement trends.
\subsection{Discretization}
\begin{figure}[htbp]
\centering
\subfloat[]{\includegraphics[width=1.7in]{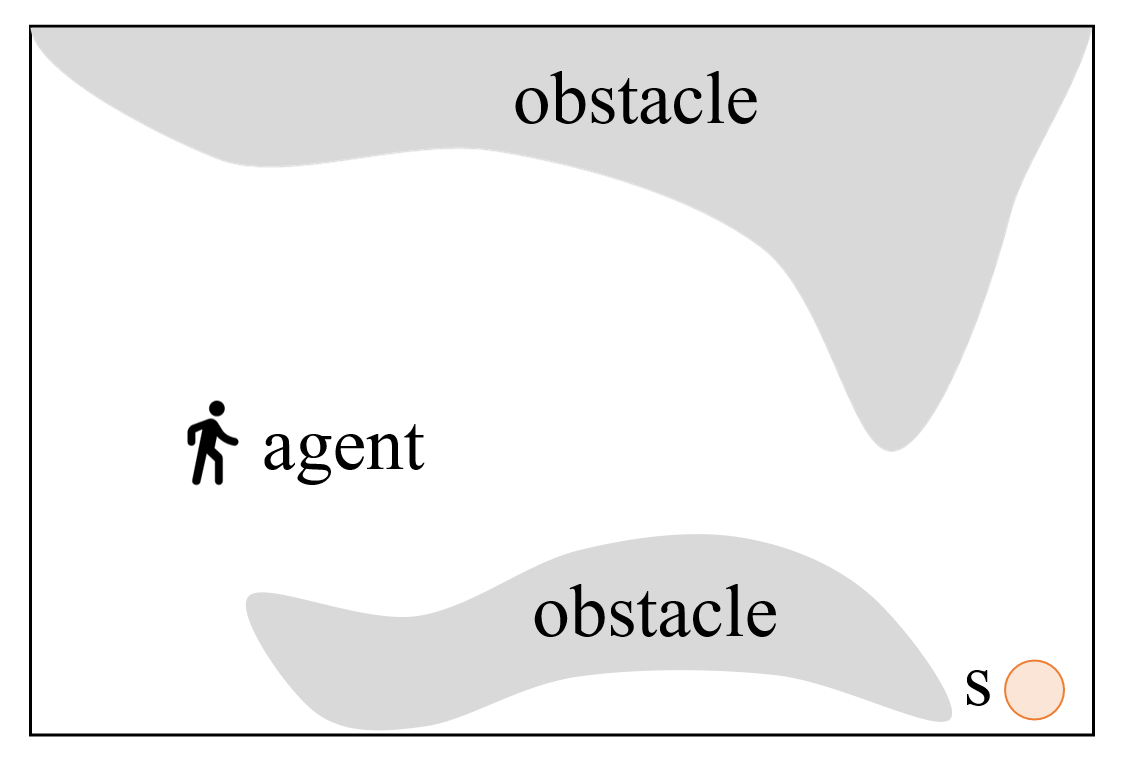}%
\label{fig_3a}}
\hfil
\subfloat[]{\includegraphics[width=1.7in]{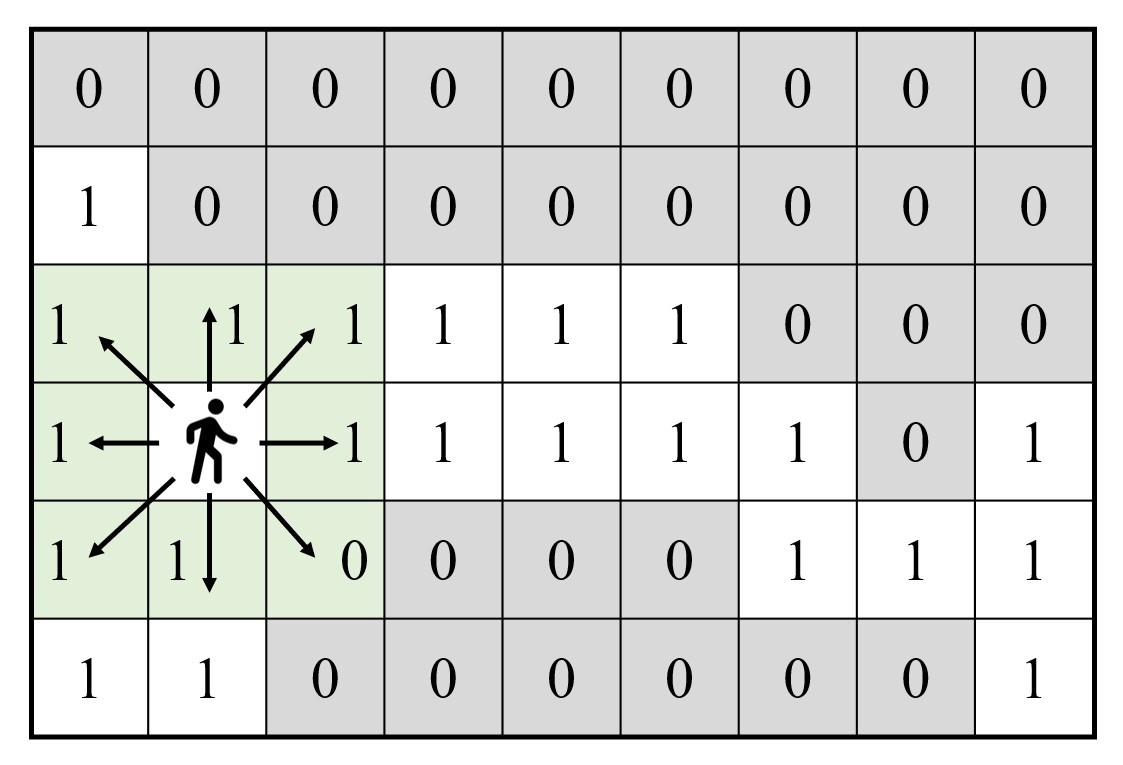}%
\label{fig_3b}}
\caption{Discretization of the environment with regular two-dimensional grids. (a) Crowd environment. (b) Discretized grids.}
\label{fig:3}
\end{figure}
The navigation potential fields span the entire available free space and can be flexibly edited to adapt user-defined simulation specifications, which are extremely suitable for guiding agent movements. At the outset, the processing and computation of a navigation potential field requires discretization of the free space in the environment. For simplicity, the space occupied by the crowd is discretized by regular two-dimensional grids, with the grid cell size slightly larger than the average diameter of a standing pedestrian (0.4 meter)~\cite{16}. As shown in Figure \ref{fig:3}, each cell is labeled as "free" or "obstacle," denoted by 1 or 0, respectively. Additionally, it is assumed that a free cell is connected to neighboring cells in eight directions.
\subsection{Basic Definition of an Navigation Potential Field}

\begin{figure}[htbp]
\centering
\subfloat[]{\includegraphics[width=1.7in]{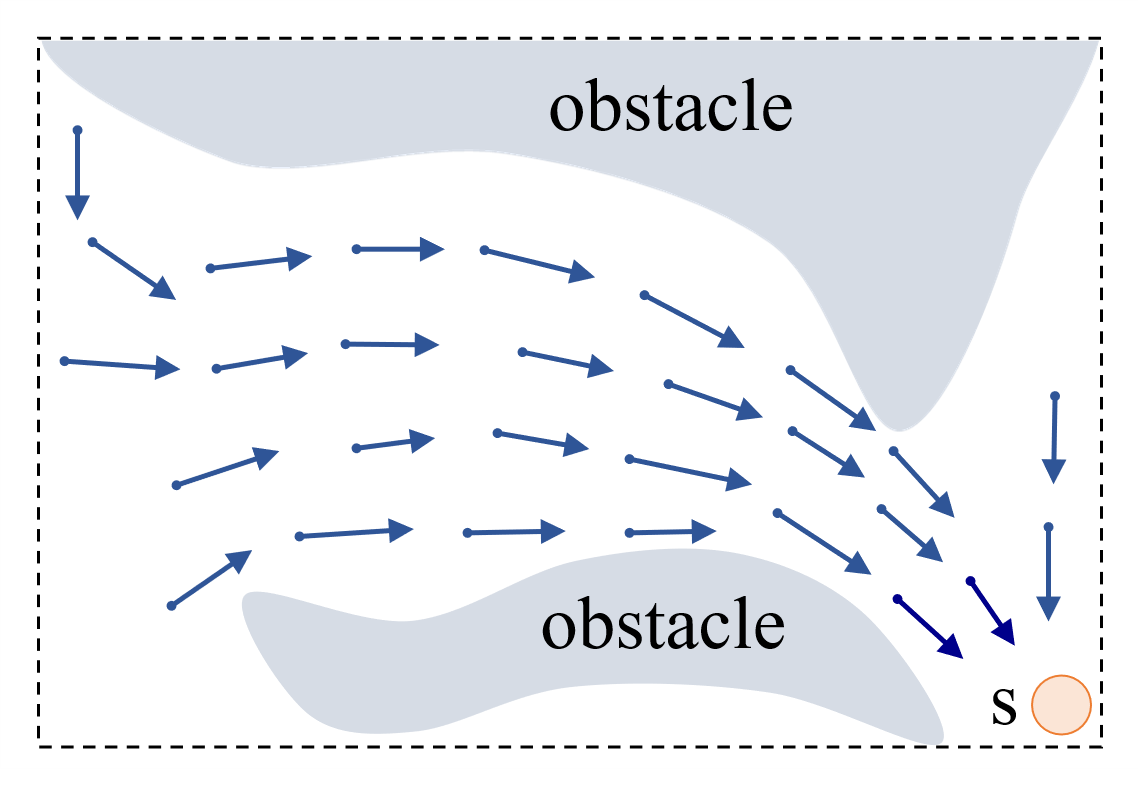}%
\label{fig_4a}}
\hfil
\subfloat[]{\includegraphics[width=1.7in]{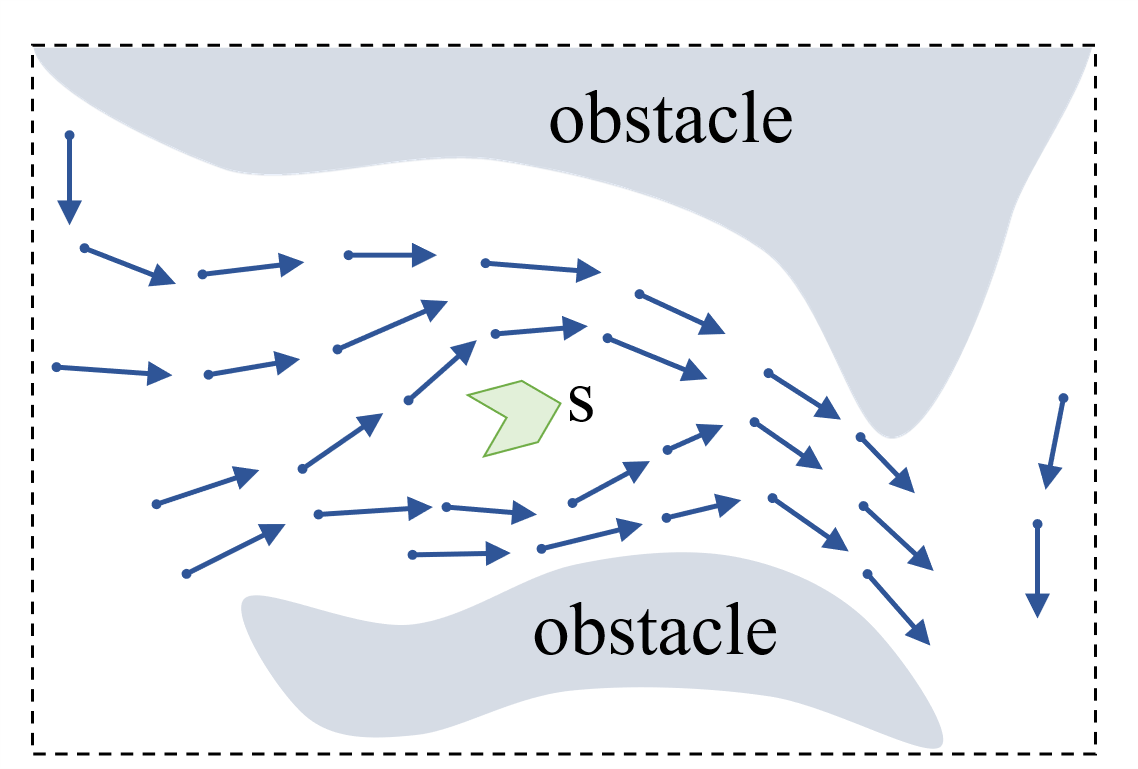}%
\label{fig_4b}}
\caption{Potential fields generated by different types of sources. (a) Source that attracts crowds. (b) Source that repels crowds.}
\label{fig:4}
\end{figure}

A single navigation potential field describes the direction that an agent will adopt near a specific source, which is denoted by $s$. The source may be other agents, obstacles, or any other environmental factors that induce a specific behavior in the agent. Some sources attract crowds (Figure \ref{fig_4a}), such as exits or destinations, while others repel them (Figure \ref{fig_4b}), like obstacles.

In Cartesian coordinate system, the effective range of $s$ is defined as region $D$, then a navigation field with source $s$ and domain $D$ is represented as a vector field $PF_{s, D}$.
\begin{equation} 
P F_{s, D}: D \rightarrow \mathbb{R}^{2}
\label{eq:PF}
\end{equation}
Each vector represents a 2D vector $P F_{s, D}\left(p\right)$ that an agent applies as its direction at position $p\in D$. After spatial discretization, the direction of $PF_{s, D}$ is transformed into direction vectors pointing toward the centers of the eight neighboring grids.

\subsection{Construction of the Navigation Potential Fields Based on PI-STGCN}
\begin{figure}[htbp]
\centering
\subfloat[]{\includegraphics[width=1.7in]{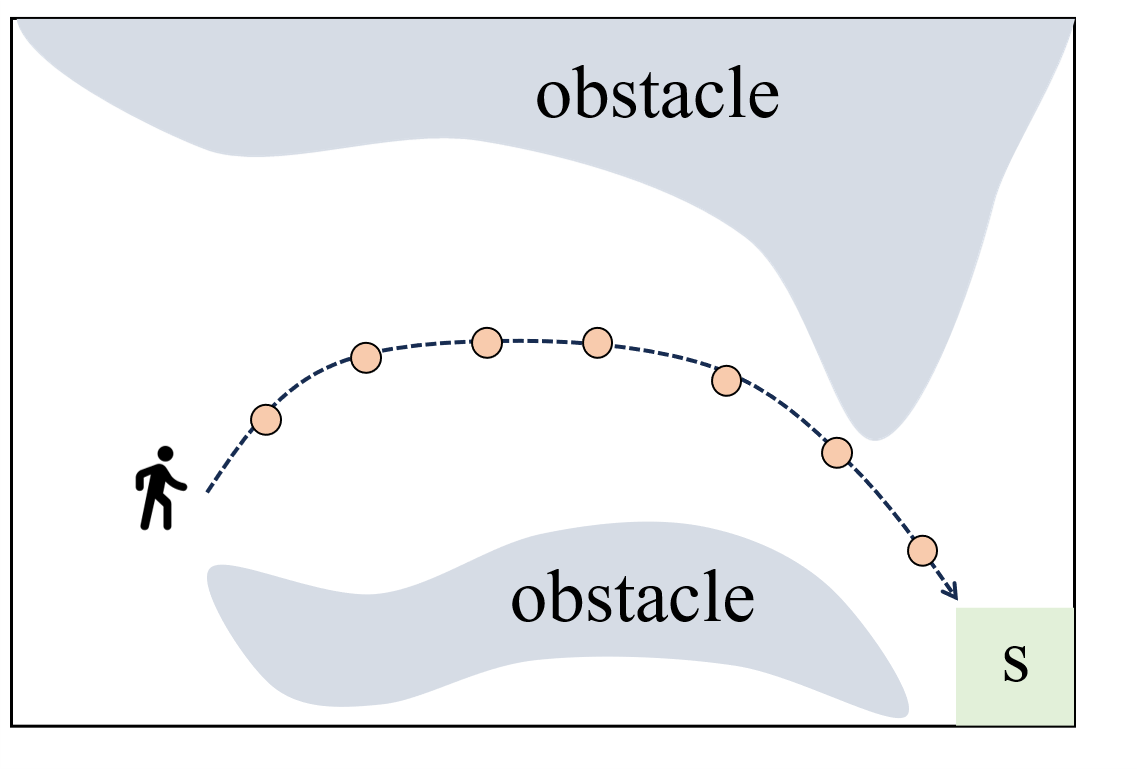}%
\label{fig_5a}}
\hfil
\subfloat[]{\includegraphics[width=1.7in]{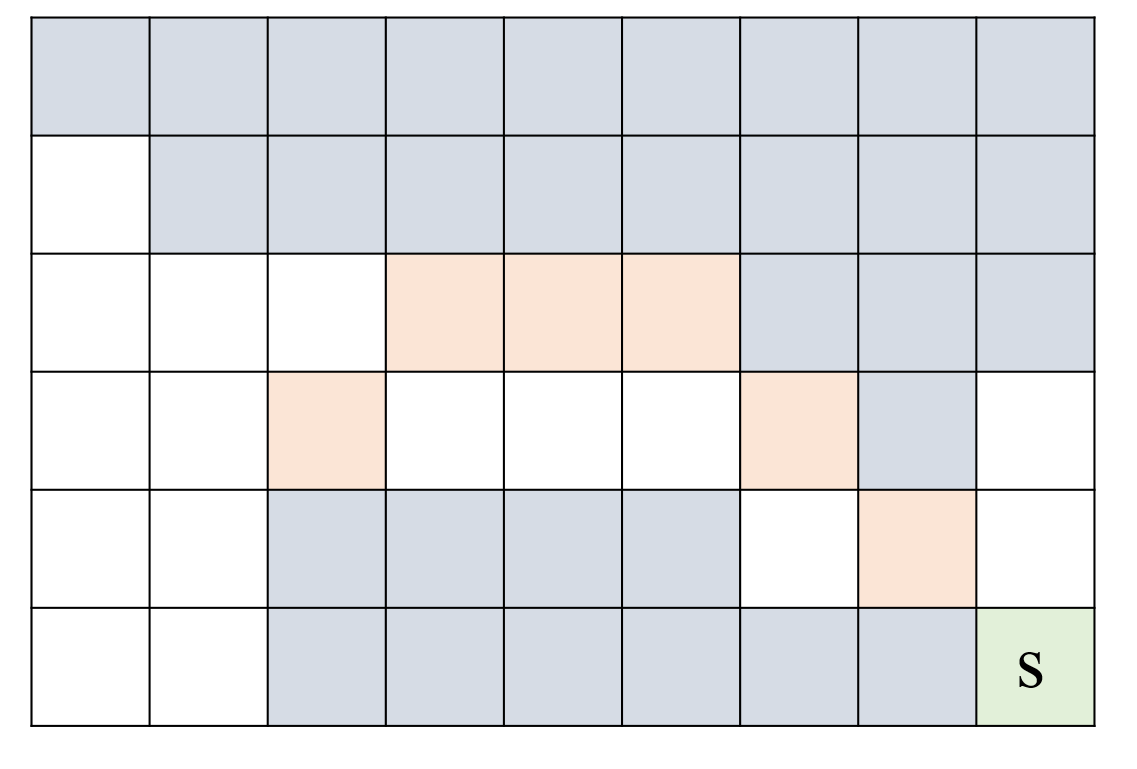}%
\label{fig_5b}}
\hfil
\subfloat[]{\includegraphics[width=1.7in]{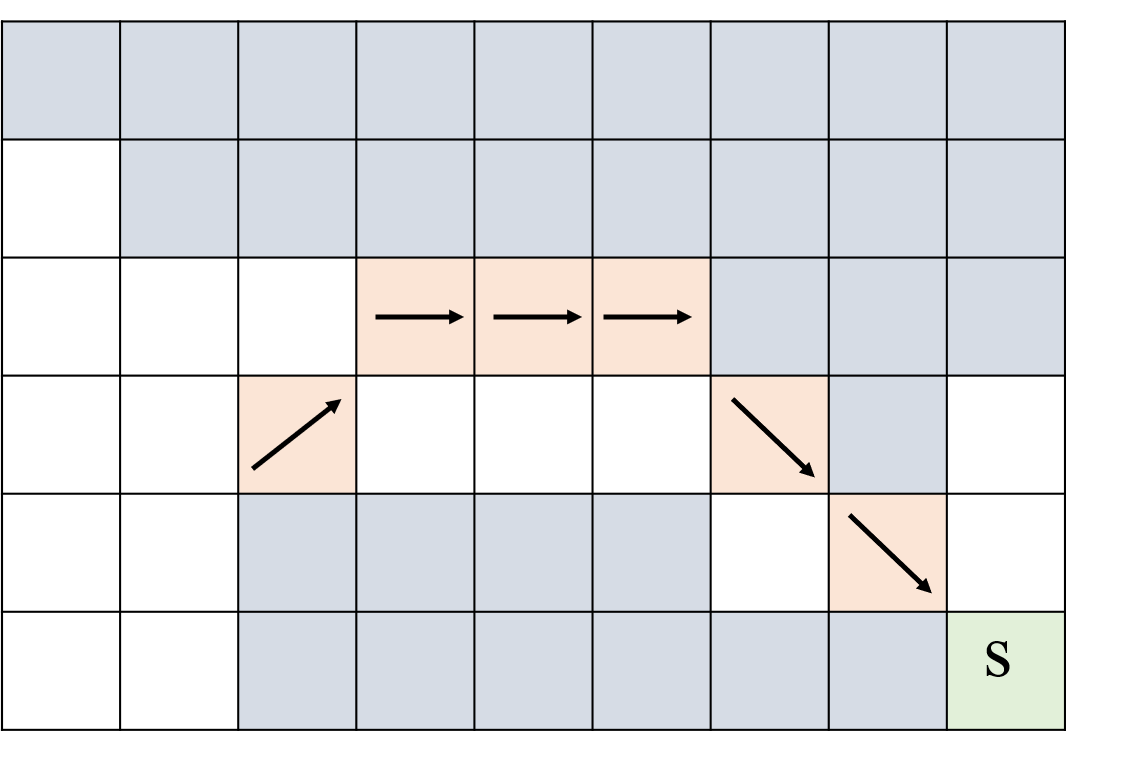}%
\label{fig_5c}}
\hfil
\subfloat[]{\includegraphics[width=1.7in]{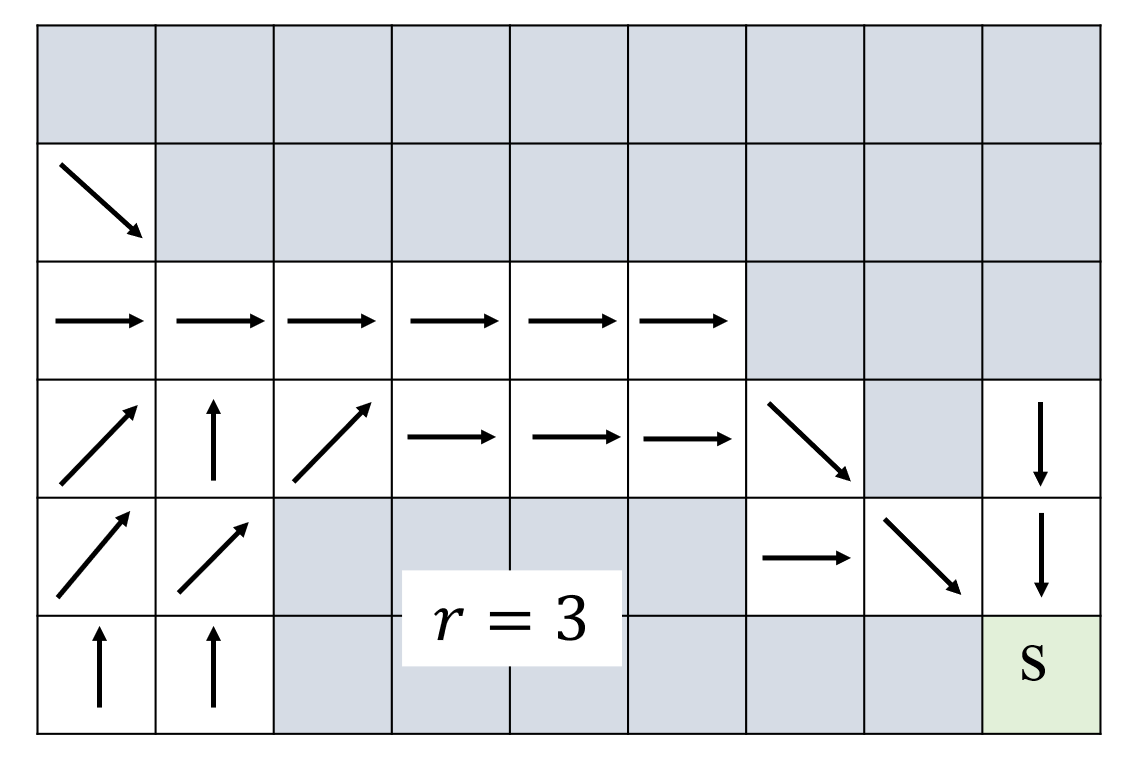}%
\label{fig_5d}}
\caption{Calculate navigation potential fields based on the predicted results of pedestrian's position distribution. (a) Sampling based on the predicted position distribution function. (b) Moving trend line. (c) Guiding direction vector. (d) Navigation potential field.}
\label{fig:5}
\end{figure}

Based on PI-STGCN's output of pedestrian position prediction, the distribution of the future positions of the agent is obtained at the current time, denoted by $\hat{S} \sim N(\hat{\mu}, \hat{\sigma}, \hat{\rho})$. For a distribution prediction sequence of length $T_{p}$, position points $\left\{S_{1}, S_{2}, \cdots S_{T_{p}}\right\}$ are generated by randomly sampling from a binary Gaussian distribution at each time point (Figure \ref{fig_5a}).
All position points of sampling describe possible movement trends of pedestrians. 
Let the target grid of a pedestrian as a source $s$, which attracts the pedestrian to move along a predetermined trend line (Figure \ref{fig_5b}).
Then, interpolations are performed on the nodes within $\left\{S_{1}, S_{2}, \cdots S_{T_{p}}\right\}$ to generate a continuous movement trajectory line formed by connecting the grids in sequence. In this phase, the A* algorithm~\cite{65} is utilized to plan the shortest path between $S_{t}$ and $S_{t+1}$, supporting fast generation of movement trend lines. $G$ is represented by Equation \eqref{eq:G}, where $g_{i}$ denotes a grid. The last grid $g_{n}$ represents the destination where the agent is moving to.

\begin{equation}
G=\left\{g_{1}, g_{2} \ldots, g_{n}\right\}
\label{eq:G}
\end{equation}
Calculating the difference between the coordinates of adjacent grids in $G$ as Equation \eqref{eq:F_gi}, the raw direction vector is obtained based on the trend line (Figure \ref{fig_5c}).
\begin{equation}
F_{g_{i}}={\left\langle g_{i+1}^{x}-g_{i}^{x}, g_{i+1}^{y}-g_{i}^{y}\right\rangle} \quad, \forall g_{i} \in G
\label{eq:F_gi}
\end{equation}

In reality, pedestrian movements exhibit randomness, and pedestrians may move to any available grid units near $G$. Taking the grid center in $G$ as the center of the navigation potential field, the radius of field domain $D$ is set to $r$ grids. Every grid $p$ within the range of $r$ is attracted by the field.
Finally, the raw direction vectors are expanded outward. Before the first expansion, only the grids in $G$ are marked. During each expansion, unmarked grids randomly select the nearest marked grid and calculate the direction vector pointing towards that marked grid, and then the unmarked grid is marked. After $r$ expansions, a pedestrian navigation potential field based on the PI-STGCN's position prediction is constructed, denoted as $F$ (Figure \ref{fig_5d}). In equation \eqref{eq:F}, $g_{i}\leftrightarrow g_{j}$ represents that $g_{i}$ is adjacent to $g_{j}$.
\begin{equation}
F=\left\{\begin{array}{ll}
F_{g_{i}}, & \text { when } g_{i} \in G \\
\left\langle g_{j}^{x}-g_{i}^{x}, g_{j}^{y}-g_{i}^{y}\right\rangle, & \text { when } g_{i} \notin G \text { and } g_{i}\leftrightarrow g_{j} \\
\text { infinite, } & \text { when } g_{i} \text { is obstacle }
\end{array}\right.
\label{eq:F}
\end{equation}

\subsection{Collision Avoidance and Pedestrian Interaction}

\begin{figure}[htbp]
\centering
\subfloat[]{\includegraphics[width=1.7in]{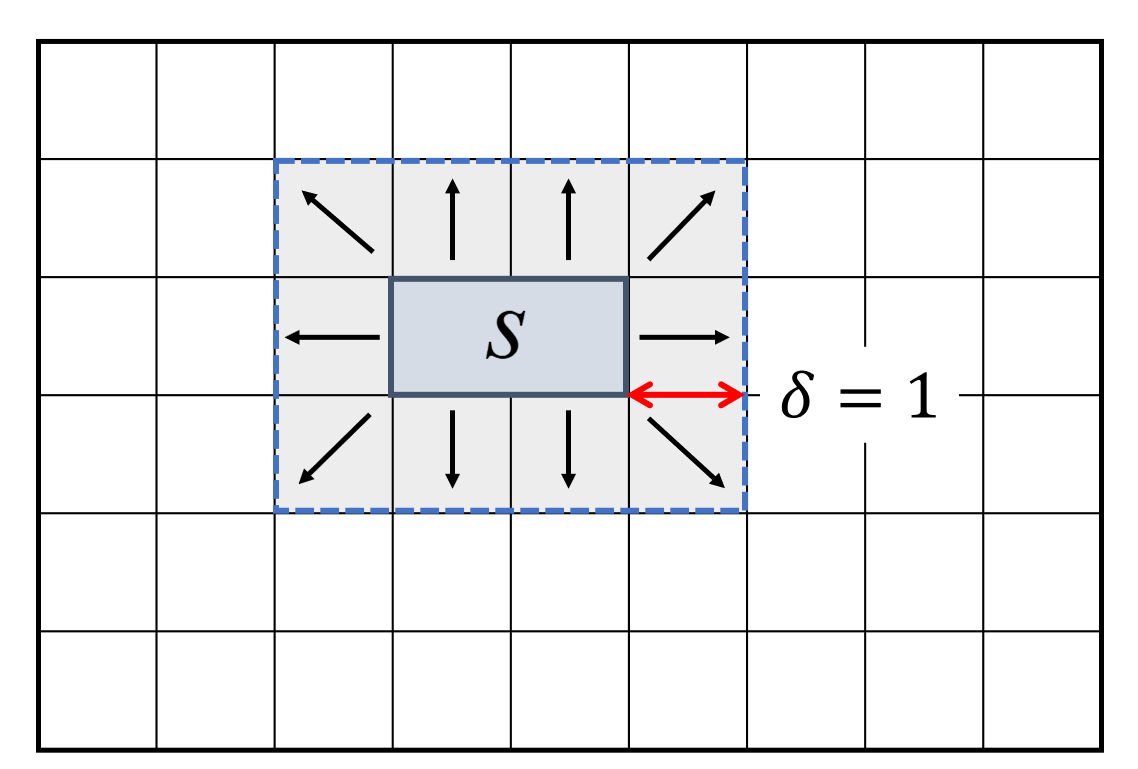}%
\label{fig_6a}}
\hfil
\subfloat[]{\includegraphics[width=1.7in]{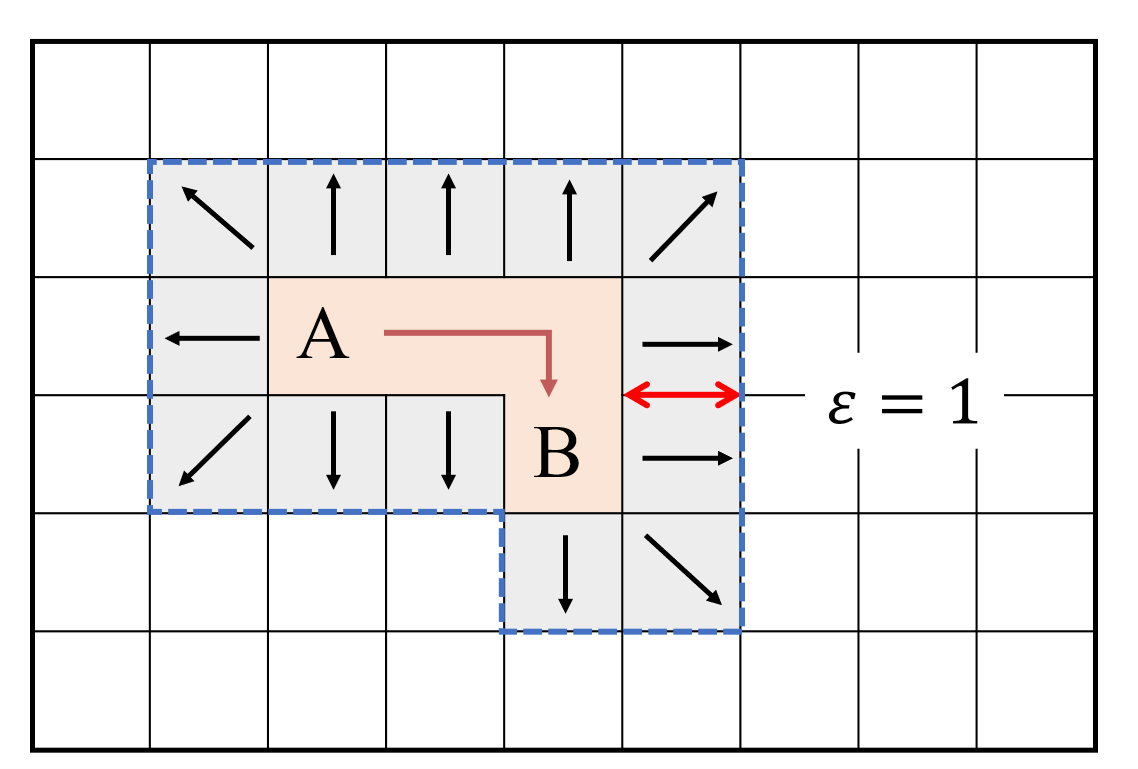}%
\label{fig_6b}}
\caption{Potential fields representing collision avoidance and pedestrian interaction. (a) Obstacle field. (b) Pedestrian field.}
\label{fig:6}
\end{figure}

In a simulated environment that uses a discrete grid representation, all obstacle objects are assumed to be regular geometric shapes or assemblies of multiple geometric shapes. The obstacle field denoted as $C$ is designed to describe the obstruction effect of obstacles on pedestrians. As shown in Figure \ref{fig_6a}, the contour lines of the obstacle objects are expanded from the edges of the outermost grid, and the obstacle field is perpendicular to the contour edges.
Let $\delta$ be the range of the obstacle field, $\lambda_{o}$ be the proportionality factor characterizing the strength of obstruction effect of obstacles, and $\vec{O}$ be the shortest vector pointing from the obstacle surface to an external position. The obstacle field at that position is:
\begin{equation}
C=\left\{\begin{array}{ll}
\lambda_{o}(\delta-|\vec{O}|) /(|\vec{O}| \cdot|\vec{O}|) \cdot \vec{O}, & \text { when } 0<|\vec{O}| \leq \delta  \\
0, & \text { when }|\vec{O}|>\delta \\
\text { infinite, } & \text { when }|\vec{O}|=0 
\end{array}\right.
\label{eq:C}
\end{equation}

Similarly, a pedestrian field is used for maintaining distances between individuals during pedestrian interactions, and it is denoted with $I$. Unlike obstacles, the positions of pedestrians are dynamically changing, so the pedestrian field excited by agent towards the surrounding crowd varies dynamically. Within a short time step, the movement of pedestrian over a small area is considered as exciting a temporary static field. If a pedestrian moves from position A to position B within time $t$, the space swept through during this movemnt is considered as the source location of the pedestrian field. 
As shown in Figure \ref{fig_6b}, let $\varepsilon$ be the effective range of the field, which represents the designated distance range of influence between pedestrians. $\lambda_{h}$ is the proportionality factor characterizing the strength of the interaction, and $\vec{H}$ is the shortest vector pointing from the space swept through by the pedestrian's movement to an external position. Similar to the method with obstacle fields, the field describing pedestrian interactions is constructed for the time period from $t-\delta t$ to $t$ by Equation \eqref{eq:I}.
\begin{equation}
I=\left\{\begin{array}{ll}
\lambda_{h}(\varepsilon-|\vec{H}|) /(|\vec{H}| \cdot|\vec{H}|) \cdot \vec{H}, & \text { when } 0<|\vec{H}| \leq \varepsilon \\
0, & \text { when }|\vec{H}|>\varepsilon \\
\text { infinite, } & \text { when }|\vec{H}|=0 
\end{array}\right.
\label{eq:I}
\end{equation}

\subsection{Applying Navigation Potential Fields During Simulation}
In simulations, to improve computational efficiency, the potential field is converted into a numerical matrix. In this manner, the navigation fields can quickly guide agents in planning paths and selecting target areas.
Based on the timeliness of the fields, the fields that influence on agents at any given time contain static fields and dynamic fields. Among them, obstacle fields are static fields constructed based on rules, while navigation potential fields and pedestrian field are dynamic fields. Adopting a numerical field representation strategy, the effect on agents at any moment is described with a global navigation potential field defined as $GF$ (Equation~\ref{eq:GF}).
\begin{equation}
GF_{t}=C+F_{t}+I_{t}
\label{eq:GF}
\end{equation}
For the navigation potential fields calculated by predicting the position distribution using PI-STGCN, it is assumed that agents tend to move from areas of high field values to areas of low field values.
The endpoint grid $g_{T}$ exhibits a minimum field value of $v_{0}$. For its eight neighboring grids, if the vector corresponding to a neighboring grid $g_{i}$ points towards the endpoint grid, then the field value $v_{0,i}$ of $g_{i}$ is given by $v_{0}+l \times\left|\vec{f}_{i}\right|$, where $l$ is the gradient scale factor set for adjacent grids.
In this paper, we disregard factors such as slopes and other elements influencing passage, assuming that $l$ between different grids is consistent. If a grid is labeled as "obstacle," its field value is set to infinity.
After calculating the field value of grid $g_{i}$, the field value $v_{0,i,j}$ of its neighboring grid $g_{j}$ is iteratively computed. Through a finite number of iterations, a navigation potential field is converted into a numerical matrix $\mathbf{M}_{\mathbf{F}}$, where each element $M_{F}^{i j}$ represents the field value corresponding to the grid at coordinates $(i,j)$. 
For obstacle fields and pedestrian fields, the field values are directly obtained by calculating the modulus of the grid vectors and then added to matrix $\mathbf{M}_{\mathbf{F}}$ to compute the global navigation potential fields. Therefore, the computation of the numerical matrix $\mathbf{M}_{\mathbf{G}}$ for a global navigation potential field is represented as Equation \eqref{eq:M_G}

\begin{equation}
\mathbf{M}_{\mathbf{G}}=\mathbf{M}_{\mathbf{F}}+\mathbf{M}_{\mathbf{C}}+\mathbf{M}_{I}
\label{eq:M_G}
\end{equation}
For any grid at coordinates $(i,j)$,
\begin{equation}
M_{F}^{i j}=\sum_{k=g_{i j}}^{g_{T}} l \times\left|\vec{f}_{k}\right|+v_{0}
\label{eq:M_F_ij}
\end{equation}

\begin{equation}
M_{C}^{i j}=\left|\overrightarrow{C_{i j}}\right|
\label{eq:M_C_ij}
\end{equation}

\begin{equation}
M_{I}^{i j}=\left|\overrightarrow{I_{i j}}\right|
\label{eq:M_I_ij}
\end{equation}

Pedestrians prefer to choose grids with smaller field values from accessible areas as their priority targets. In simulation, agents can quickly select the target grid based on the field values of neighboring grids from their global navigation potential fields. When an agent completes a movement and its position is updated, the pedestrian field at time $t$ automatically disappears, while a new pedestrian field is generated at time $t+\Delta t$. The data-driven period is denoted as $T_{d}$, and $1 \leq T_{d} \leq T_{p}$. After $T_{p}$ time steps in simulation, the navigation potential fields are updated based on new historical movement data in a data-driven mode through the PI-STGCN. Algorithm \ref{alg:n_p_f} describes the computing and updating process of the navigation potential fields during the simulation.

\begin{algorithm}[htbp]
    \caption{The computing and updating process of the navigation potential fields.}
    \label{alg:n_p_f} 
    \KwIn{Initial crowd of agents, \\ \hspace{2.8em} Discretized environment, \\ \hspace{2.8em} The data-driven period $T_{d}$.}
    \KwOut{Crowd dynamics}
    
    Initialize navigation potential field\;
    Calculate global navigation field by Equation \eqref{eq:GF}\;
    \While{not at end of simulation}{
         $t+=\Delta t$\; 
         \For{\textit{each agent in crowds}}{
             choose the movement route based on $\mathbf{M}_{\mathbf{G}}$\;
             Execute the movement action\;
             \eIf{Arrival at the endpoint grid}{
                 Remove the agent from the crowd\;
                 break\;
             }{} 
             Update the position of the agent\;
         }
         Update the pedestrian field\;
         \eIf{$t$ pass an interval of $T_{d}$}{
             Predict the movement trends by PI-STGCN\;
         }{}
         Update the global navigation potential fields\;
    }
\end{algorithm}
\section{Experiments}
\subsection{Datasets and Setup}
We evaluate our proposed crowd simulation framework on two public large-scale real-world datasets, including the GC dataset~\cite{55} and the UCY dataset~\cite{56}.
The GC dataset contains pedestrian data from two scenarios: ETH and Hotel, while the UCY dataset includes pedestrian data from three scenarios: Univ, Zara1, and Zara2. These datasets record pedestrian trajectories in real-world environments, consisting of thousands of trajectories from over 1,500 individuals across five scenarios. Different scenarios exhibit variations in crowd size, trajectory distribution, and social behaviors. 
Based on the aforementioned datasets, pedestrian information with a sequence length compatible with the PI-STGCN's input is extracted from real crowds to create agents for crowd movement simulation experiments. The extracted simulation input data for the agents is shown in Table \ref{tab:II}. In the UNIV scenario, pedestrians are the most concentrated, with the slowest average movement velocity, while in the ETH scenario, pedestrians exhibit the maximal average velocity.
\begin{table}[!htbp]
\centering
\caption{Agent information extracted from two real-word datasets across five scenarios.}
\label{tab:II}
\renewcommand{\arraystretch}{1.5}
\begin{tabular}{@{}lccccc@{}}
\hline
\hline
                       & ETH   & HOTEL & UNIV  & ZARA1 & ZARA2 \\ \hline
The number of agents   & 255   & 112   & 364   & 139   & 185   \\
Video length (seconds) & 578.8 & 440.4 & 215.6 & 346.8 & 420   \\
Average velocity (m/s)      & 1.625 & 1.01  & 0.89  & 1.24  & 1.19  \\ 
\hline
\hline
\end{tabular}
\end{table}

To enhance the generalization performance of our data-driven framework by leveraging diverse data distributions, cross-validation is employed to divide the datasets across five scenarios into training and testing sets. In this setup, one scenario serves as the validation and testing set while the remaining scenarios are used for training. To maintain consistency with the time intervals of the input data sequences used in the PI-STGCN, the simulation time step $\Delta t$ is set to 0.4 seconds. Based on previous research~\cite{32}, the PI-STGCN is configured to use 8 time steps of historical data to predict the movement trends for the next 12 time steps, i.e., $T_{p}=12$. Since the minimum spacing between pedestrians in the real crowd data is less than the grid edge length of 0.4 meters, the boundary size for both the obstacle field and the pedestrian field is set to 1 grid.

\subsection{Performance Analysis}
\begin{figure*}[htbp]
	\centering
	\includegraphics[width=1\linewidth]{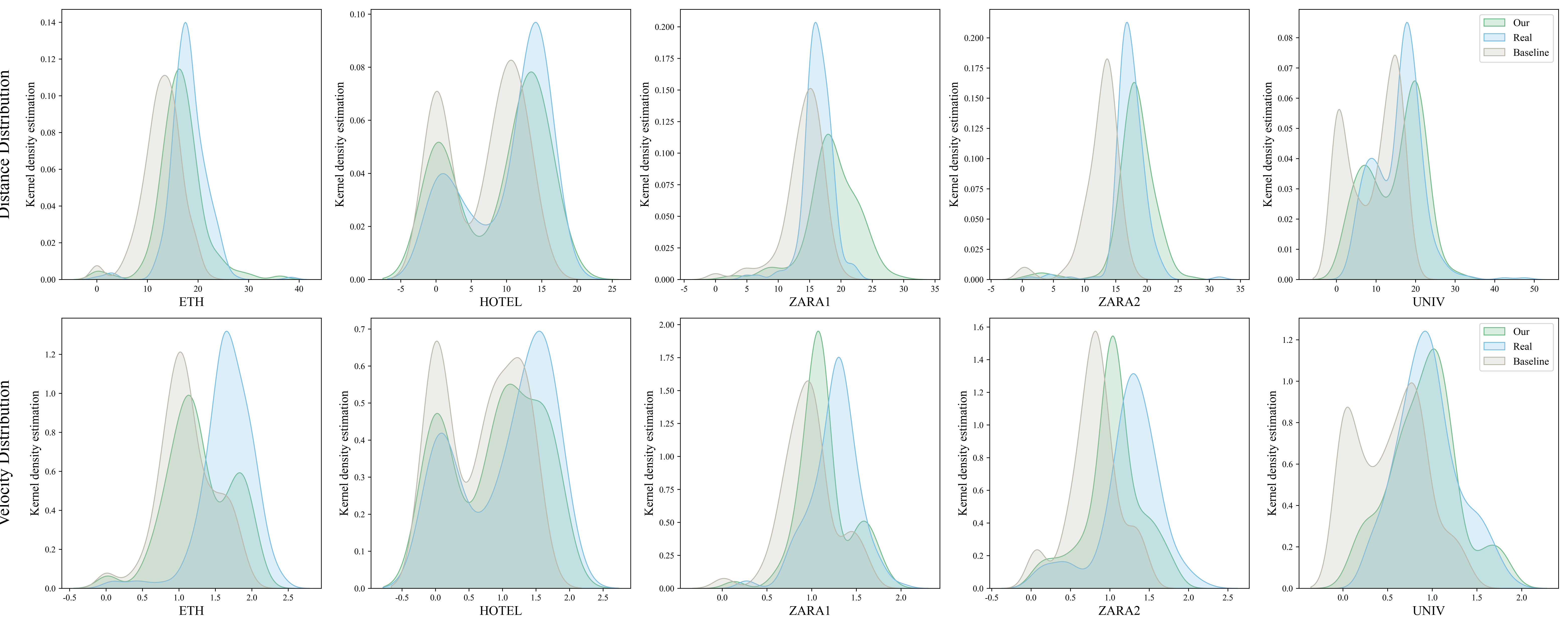}
        \centering
	\caption{The kernel density distributions of movement distances and average velocities of agents from the simulation results of the five scenarios.}
	\label{fig:7}
\end{figure*}

The accuracy of the simulation can be evaluated by quantitatively comparing the simulated crowd trajectories with the real trajectories. As the lack of constraints on the interactions between the environment and pedestrians, purely data-driven methods are not suitable for direct application in long-term simulations and may not facilitate quantitative comparison. It is found that generating simulated trajectories solely based on deep learning over a long term lead to results that violate real-world conditions, such as pedestrians entering obstacle areas and occupying the same space. Therefore, a rule-based model is selected as the baseline for performance comparison, while the advantages of our method compared to purely data-driven methods will be discussed in subsequent trajectory analysis.
The rule-based model considers only the direction from the current position to the destination and follows the shortest path to determine the next grid step, lacking any real-time movement trend information.
The same evaluation metrics as those in~\cite{53} are used for assessment, comparing the statistical results of the simulated crowd's average velocity and movement distance with the distribution found in the real data. 
By analyzing the results obtained from five experimental scenarios, the kernel density distributions of the total movement distance and average movement velocity for all agents are presented in Figure \ref{fig:7}.

\begin{figure*}[htbp]
	\centering
	\includegraphics[width=1\linewidth]{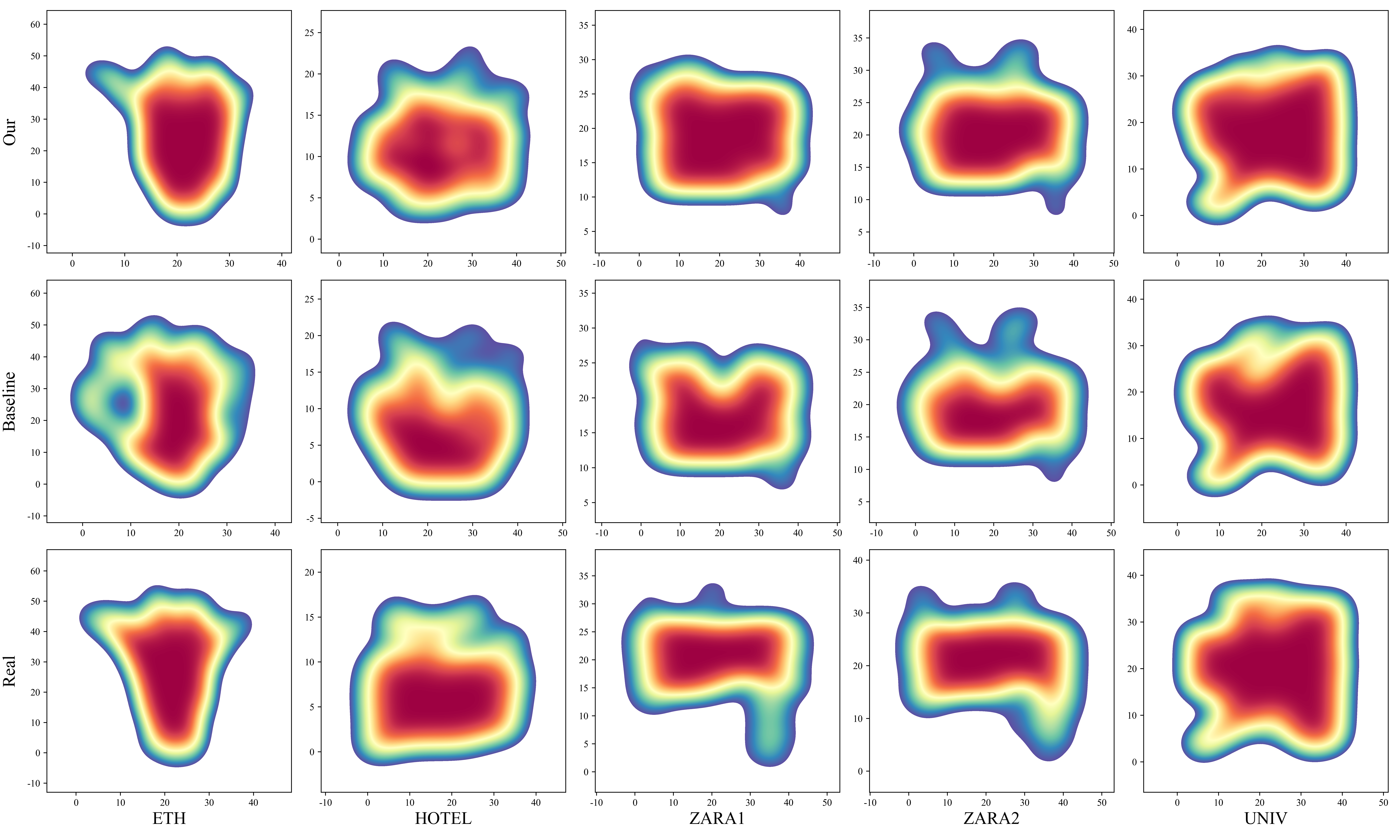}
        \centering
	\caption{The heat distribution constructed from the trajectories of agents based on the simulation results of five scenarios. Areas with higher pedestrian densities in increasingly red tones on the heatmap.}
	\label{fig:8}
\end{figure*}
The crowd movements simulated based on our proposed framework closely resemble real-world scenarios. Conversely, the baseline struggles to accurately simulate real pedestrian behaviors by prioritizing the shortest path.
Consequently, the total movement distances simulated by the baseline are less than the truth value. As the presence of obstacles, pedestrian interactions and other factors in real crowd movements, pedestrians do not consistently move in a straight line or following the shortest route to their destination. Instead, they exhibit certain angular deviations and take circuitous paths sometimes. In our framework, the PI-STGCN effectively captures pedestrian movement trends, resulting in simulated crowd movement distances and velocity distributions that more accurately correspond to real-world scenarios. Specifically, in the densely populated UNIV scenario, the distance distribution curve of the real data reveals a significantly higher peak on the right side compared to the left. In contrast, the distribution produced by the rule-based baseline simulation shows a distinct bimodal shape, with the heights of the two peaks being relatively close together. This diverges considerably from the true distribution, whereas our simulation results are more reasonable.

To further validate the accuracy of the crowd simulation, average distance error (ADE) is adopted as an evaluation metric to compare the microscopic-level simulation trajectories of all agents with those in the real data. Let $P_{t}^{n}$ represent the true position of an agent $n$ at time $t$, while $\tilde{P}_{t}^{n}$ denotes the simulated position. The definition of ADE is given in Equation \eqref{eq:ade}.
\begin{equation}
\mathrm{ADE}=\frac{\sum_{n \in N} \sum_{t \in T_{p}}\left\|\tilde{P}_{t}^{n}-P_{t}^{n}\right\|_{2}}{N \times T_{p}}
\label{eq:ade}
\end{equation}
Table \ref{tab:III} presents the average ADE of all agents in the simulation results across the five scenarios. Except for the HOTEL scenario, where the ADE of our model's simulated results is slightly higher than the baseline, ADE values in the other four scenarios are all below the baseline. Overall, the simulation that integrates data-driven methods shows a 4\% improvement in performance compared to purely rule-based simulation. 
\begin{table}[htbp]
\centering
\caption{Comparison of ADE between trajectories generated by different methods and real-world crowd data.}
\label{tab:III}
\renewcommand{\arraystretch}{1.5}
\begin{tabular}{@{}lcccccc@{}}
\hline
\hline
         & ETH    & HOTEL   & ZARA1  & ZARA2  & UNIV   & AVG    \\ \hline
Our      & 8.93   & 6.27    & 7.98   & 8.42   & 6.18   & 7.556  \\
Baseline & 9.53   & 6.22    & 8.52   & 8.8    & 6.43   & 7.9    \\
         & \underline{\textbf{6.30\%}} & -0.80\% & \underline{\textbf{ 6.34\%}} & \underline{\textbf{4.32\%}} & \underline{\textbf{3.89\%}} & \underline{\textbf{4.01\%}} \\ \hline \hline
\end{tabular}
\end{table}
The reasons why our crowd simulation framework underperformed compared to the rule-based model are analysed in the HOTEL scenario. These five real-world crowd scenarios are characterized by their limited spatial range and relatively simple environmental obstacles. Generally, pedestrians tend to move in a straight line toward their destinations, with local deviations and stops not significantly straying from the intended path. Consequently, the calculated ADE values show little difference from the baseline, remaining within 1 meter. Particularly in the HOTEL scenario, where there are the fewest obstacles and the lowest crowd density, pedestrians tend to move in a straight line along the shortest path. Data-driven simulations may result in a slight increase in ADE values due to the excessive consideration of complex spatio-temporal dependencies within the neural network. Overall, methods that integrate PIML with physical models are more suitable for crowd simulation in complex scenarios. The analysis of specific simulation trajectories in the next section will further support our viewpoint.
\subsection{Trajectory Analysis}
\begin{figure}[!ht]
\centering
\subfloat[]{\includegraphics[width=3in]{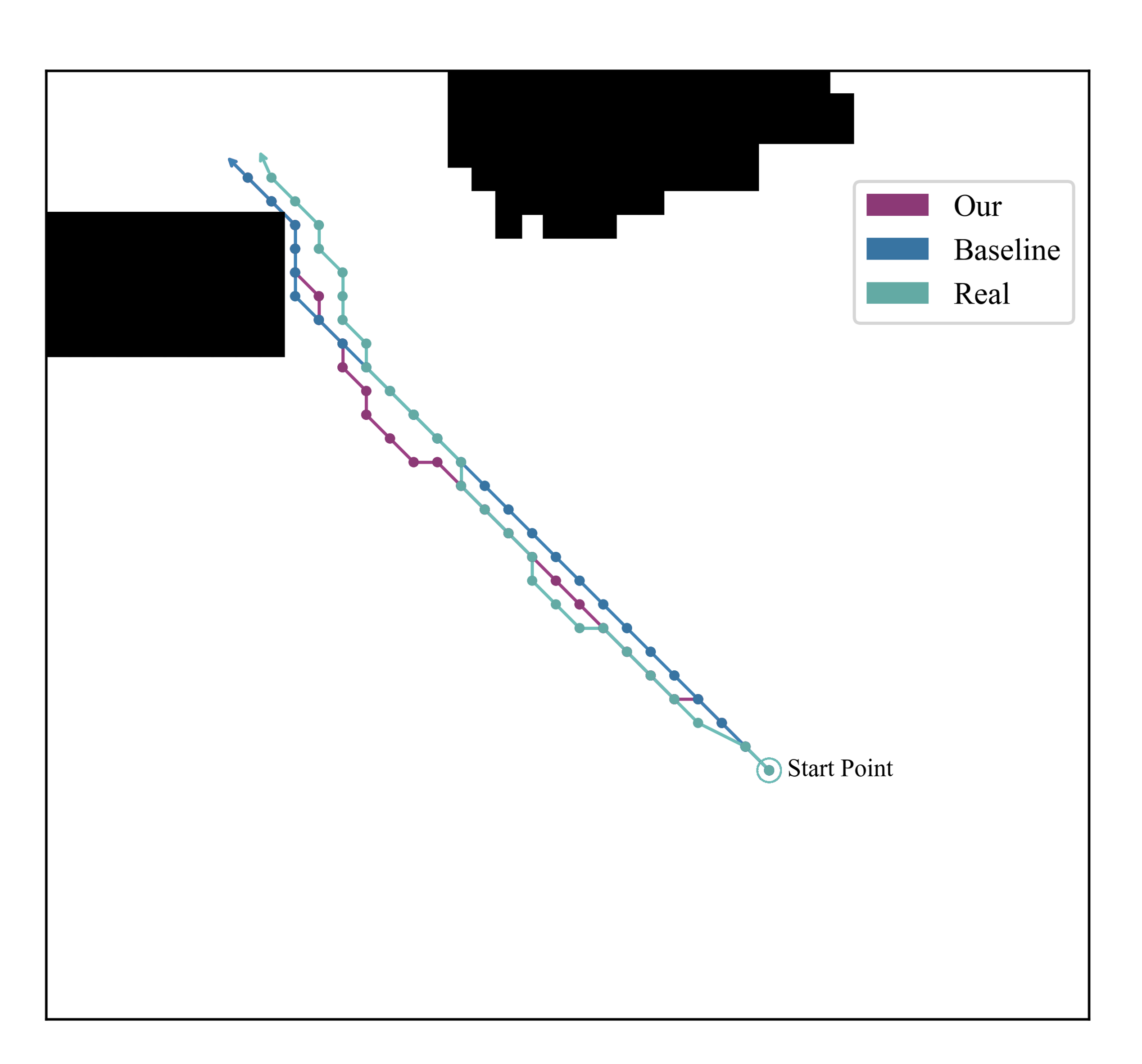}%
\label{fig_9a}}
\hfil
\subfloat[]{\includegraphics[width=3in]{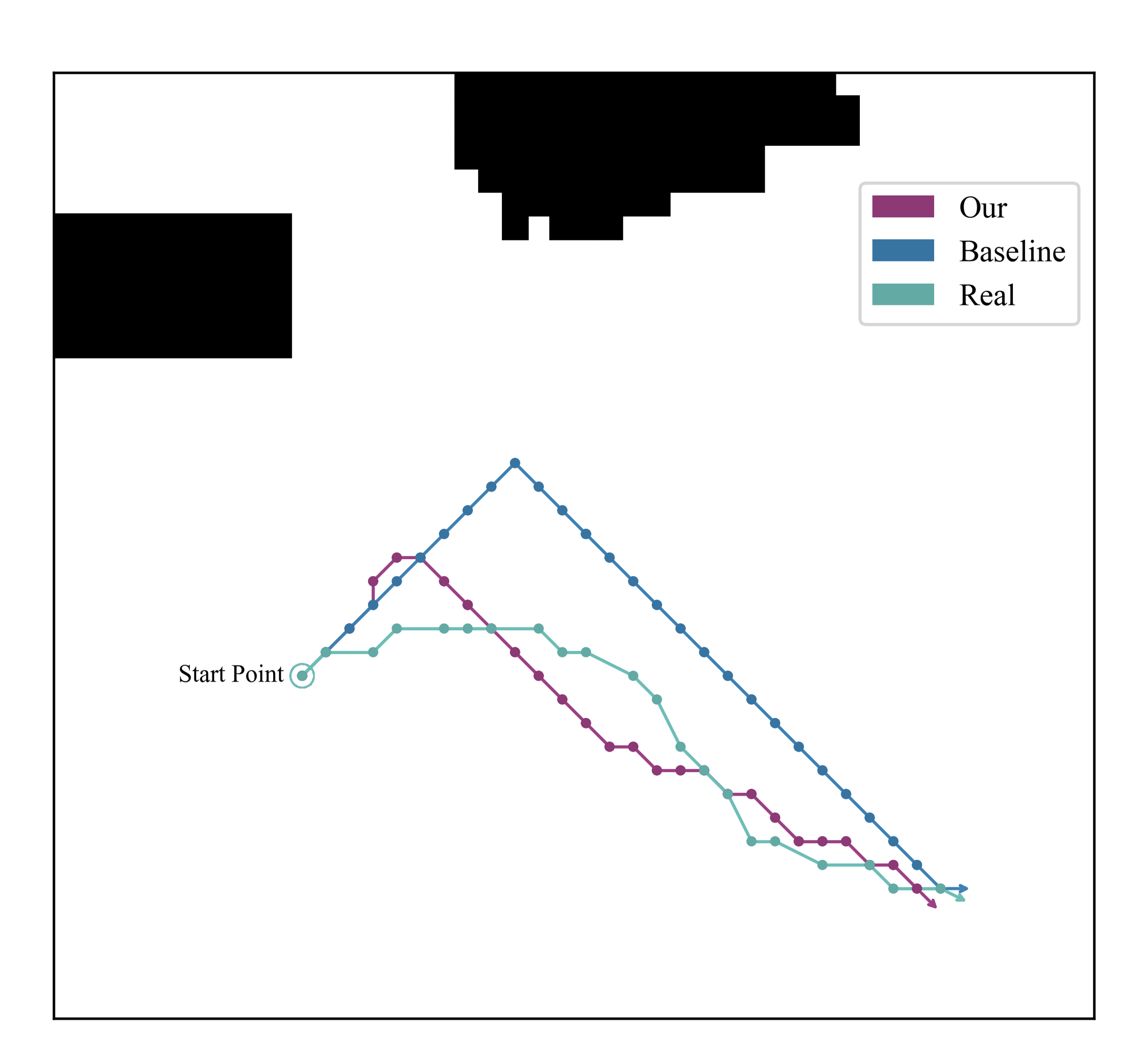}%
\label{fig_9b}}
\caption{Simulated movement trajectories of two agents in the UNIV scenario. (a) Simulated trajectory of Agent 1. (b) Simulated trajectory of Agent 2.}
\label{fig:9}
\end{figure}
The simulated crowd trajectories are visually represented, with the heatmap constructed based on the positions of agents within the trajectories. The heatmap reflects the concentration of crowd activity in different areas within the scene; regions with higher heat values indicate a greater number of pedestrians passing through. Figure \ref{fig:8} illustrates the heatmaps of crowd trajectories simulated in five different scenarios alongside the heatmaps extracted from real-world datasets. From the perspective of trajectories, both methods generate relatively realistic crowd movements.  Nevertheless, there are certain differences in the crowd distribution produced by the two methods. From a visual standpoint, the movements simulated based on our methods are more similar to the real crowd trajectory distributions across all scenarios except for HOTEL. This similarity is particularly distinct in the ETH, ZARA2, and UNIV scenarios.
The Jaccard similarity coefficient~\cite{57} is calculated to evaluate the similarity between the different levels of heatmap areas generated by the two methods and the actual values. The results presented in Table \ref{tab:IV} demonstrate that our method surpasses the baseline in average similarity, indicating greater accuracy in reproducing the movement trends and spatial distribution of real crowds. 
The trajectory analysis further confirms the effectiveness and sophistication of our proposed crowd simulation framework. Certain complex pedestrian behaviors are difficult to represent using rule-based models, while the combination of data driven and physical modeling demonstrates superior performance in capturing trajectory characteristics compared to conventional purely physical models.
\begin{table}[htbp]
\centering
\caption{Comparison of the Jaccard similarity between simulated crowd heatmaps by different methods and the real crowd heatmap.}
\label{tab:IV}
\renewcommand{\arraystretch}{1.5}
\begin{tabular}{@{}lcccccc@{}}
\hline
\hline
         & ETH   & HOTEL & ZARA1 & ZARA2 & UNIV  & AVE    \\ \hline
Our      & 0.754 & 0.37  & 0.646 & 0.658 & 0.768 & 0.6392 \\
Baseline & \underline{\textbf{0.552}} & 0.495 & \underline{\textbf{0.545}} & \underline{\textbf{0.563}} & \underline{\textbf{0.73}}  & \underline{\textbf{0.577}}  \\ \hline \hline
\end{tabular}
\end{table}

Compared to the purely deep learning methods, the crowd movements simulated based on our framework exhibit stronger adaptability to complex scenarios and possess greater interpretability. To demonstrate the advantages of our framework in this regard, the complete movement trajectories and navigation potential fields of the simulated crowd are further analysed. 
Taking the densely populated UNIV scenario as an example, the trajectories of two selected agents are plotted, along with the contour maps of their navigation potential fields at the beginning of movement, in the middle of movement, and as they approach the endpoint. As illustrated in Figure \ref{fig_9a}, the actual movement trajectory of Agent 1 predominantly follows a straight line. However, as it approaches the obstacle in the upper-left corner, the trajectory locally deflects towards the opposite side of the obstacle, resulting in some slight fluctuations. In contrast, the movement simulated by the rule-based model fails to capture this characteristic, maintaining a nearly straight trajectory with poor heterogeneity. For Agent 2, the baseline is unable to accurately capture the timing of deviations during pedestrian movement, leading to considerable discrepancies with the real trajectory, while our method effectively predicts the agent's movement trends.

\begin{figure*}[htbp]
\centering
\subfloat[]{\includegraphics[width=7in]{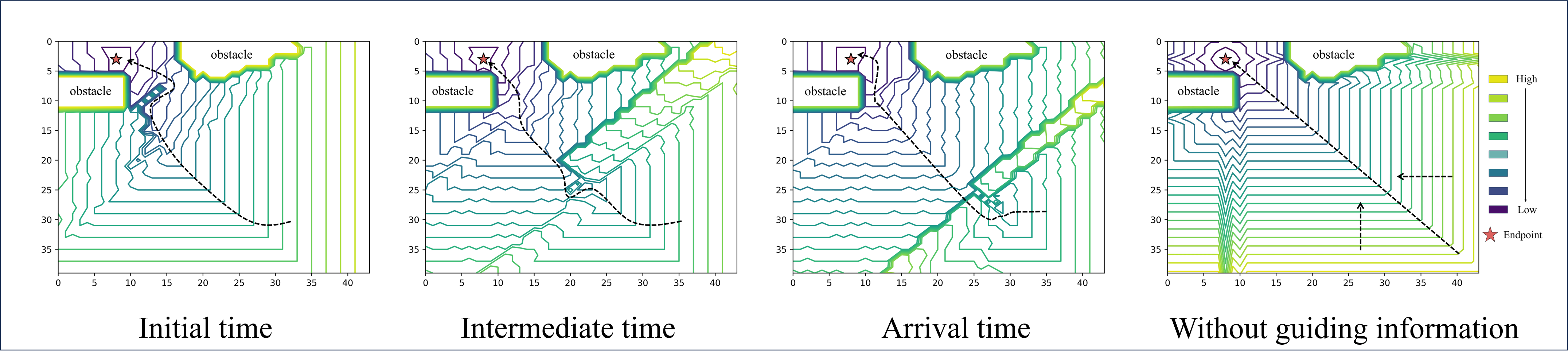}%
\label{fig_10a}}
\hfil
\subfloat[]{\includegraphics[width=7in]{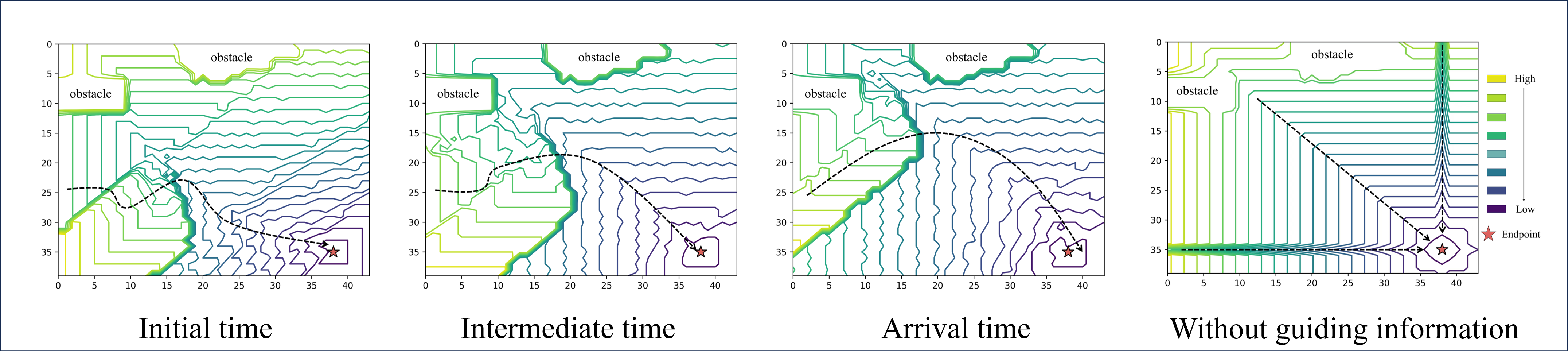}%
\label{fig_10b}}
\caption{The contour map of the navigation potential fields during the movement of Agent 1 and Agent 2. The black dashed lines indicate the approximate movement trend in the direction of decreasing gradient along the contour lines. (a) The navigation potential fields of Agent 1. (b) The navigation potential fields of Agent 2.}
\label{fig:10}
\end{figure*}

When agents entering the obstacle area, the physical constraints of collision avoidance in the navigation potential fields trigger the agents to turn and avoid obstacles. However, directly using deep learning methods to generate movement trajectories may result in routes that pass through obstacles in this scenario, which is incorrect and fails to explain the generated movement trajectories. The navigation potential fields in Figure \ref{fig:10} further demonstrate how our method makes the crowd simulation more aligned with real-world situations. In the condition where agents are driven by shortest-path and linear motion rules without guiding information, the contour lines of the navigation potential fields appear almost regular and uniform. On the contrary, the real-time injection of movement trend information predicted by PI-STGCN corrects the original navigation fields in our proposed crowd simulation framework. As shown in Figure \ref{fig:10}, Agent 1 moves toward the endpoint under the guidance of the direction perpendicular to the contour lines of it's navigation potential field. The navigation potential fields dynamically adjust and change as the simulation time progresses, providing guidance for the agents. This enhances the stability and interpretability of crowd simulation, making our framework more adaptable to long-term simulations in complex scenarios.

\subsection{The Data-Driven Period}
\begin{figure}[!ht]
	\centering
	\includegraphics[width=1\linewidth]{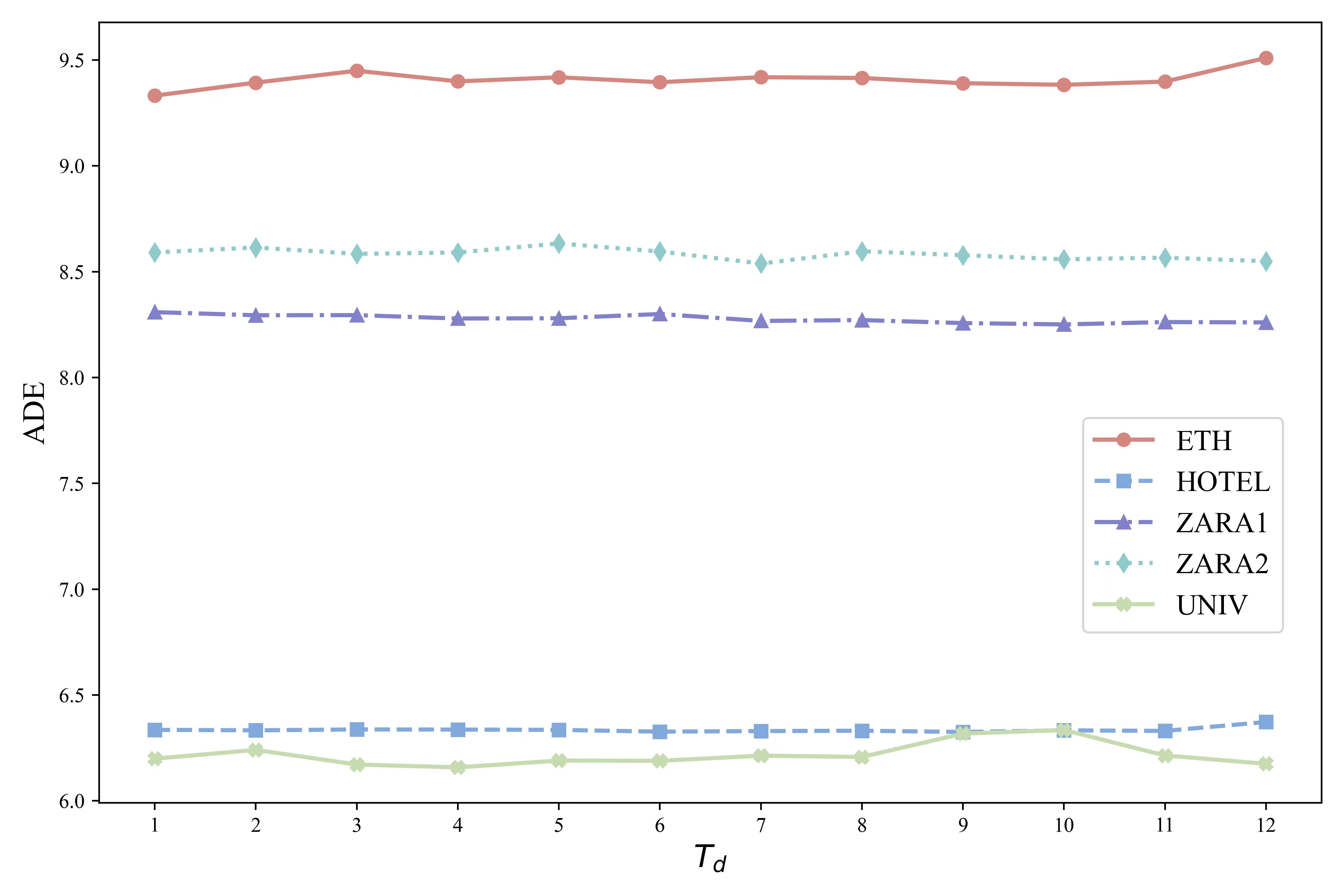}
        \centering
	\caption{Comparison of simulated ADE under different data-driven periods across five scenarios.}
	\label{fig:11}
\end{figure}

\vspace{-2pt} 
In real-time crowd simulation calculations, the update period $T_{d}$ for navigation potential fields reflects data-driven strength. If $T_{d}$ is large, it indicates insufficient data injection. In environment where crowd movement changes rapidly and exhibits significant randomness, a delayed update of navigation potential fields may occur. Consequently, outdated navigation potential fields no longer reflect the agent's current movement trend, which could lead to an increase in simulation errors. 
Exploratory experiments on data-driven period are implemented across five scenarios under various $T_{d}$ conditions, as illustrated in Figure \ref{fig:11}. The experimental results demonstrate that the update period of navigation potential fields affects the accuracy of the simulated trajectories, with ADE values exhibiting fluctuations as $T_{d}$ changes. Although the numerical differences in ADE are minor, our findings suggest that a smaller $T_{d}$ does not automatically yield a lower ADE. Instead, the lowest ADE is achieved at a moderate update period.
Considering the real pedestrian movements across the five scenarios, we hypothesize that this phenomenon arises from the relative stability of pedestrian behaviors in limited-scale environments, where individuals are more likely to maintain their movement trends over extended periods. 
When the data-driven period is too short, excessively frequent updates to the navigation potential fields lead to substantial fluctuations in the predictions output by the neural network. In such simulations, agents frequently change their movement directions, consequently leading to an increase in errors. The curves depicted in Figure \ref{fig:11} indicate that the ADE of the crowd in the relatively dense UNIV scenario and the complex ETH scenario are more sensitive to $T_{d}$, consistent with our prior analysis. Thus, in the context of practical crowd simulation tasks, particularly in small-scale environments with simple pedestrian movement patterns, it is advisable to set the data-driven period at a moderate level. This strategy not only heightens the stability of simulation but also helps conserve computational resources and improve overall computational efficiency.

\subsection{Disscussion}
In this section, we conclude the research findings and discuss some drawbacks of this work.

\textbf{\textit{Findings.}} In existing stduies of crowd simulation, rule-based models exhibit poor fidelity and heterogeneity, while models solely based on deep learning methods are limited in their generalization and interpretability. To address these challenges, a crowd simulation framework that integrates PIML with navigation potential fields is proposed. Based on this framework, more accurate and realistic simulations are successfully achieved compared to rule-based models, with an average reduction of 4\% in simulation trajectory errors on two real-world crowd datasets across five scenarios. Moreover, through the analysis of agent trajectories and the navigation potential fields in simulation, our framework demonstrates greater stability and interpretability compared to methods that directly generate simulation trajectories using deep learning methods. Additionally, we find that the data-driven period has a two-sided impact on the accuracy of simulations. These findings demonstrate the feasibility of combining traditional physical modeling methods with the latest advancements in machine learning to enhance the fidelity and applicability of crowd simulations.


\textbf{\textit{Drawbacks.}} However, there remain some drawbacks in the present work. On one hand, in predicting pedestrian movement trends, the PI-STGCN only considers the physical constraints of velocity-displacement and is limited by the assumption of uniformly accelerated motion. As the complexity of the environment increases, pedestrian movements may no longer conform to the assumed kinematic equations, making it necessary to redesign the physical loss function and retrain the neural network. Moreover, the training difficulty of the PI-STGCN will also increase.
On the other hand, the psychological effects of pedestrians are not considered in the physical model, leaving significant room for improvement in both interpretability and accuracy. The model of navigation potential fields only considers avoidance of obstacles and the maintenance of interpersonal distances among pedestrians. However, in real-world scenarios, especially when pedestrian activities are more dynamic, psychological factors have a significant influence on movement. It is unlikely that pedestrians will move strictly according to the predefined navigation fields. As a result, some simulated pedestrians may show considerable discrepancies from real behaviors when turning and stopping, and it is difficult to interpret.

\section{Conclusion}
In this study, a data-driven crowd simulation framework that integrates PIML with navigation potential fields is proposed. Specifically, the PI-STGCN is constructed to provide accurate predictions of pedestrian movement trends, and a computational algorithm of navigation potential fields is designed to support real-time path planning for agents. The effectiveness of our framework is validated by extensive experiments on two real-world datasets across five scenarios by comparing it with the rule-based model and methods based solely on deep learning. Research findings reveal that the proposed framework advances the fidelity and accuracy of crowd simulations while also exhibiting higher adaptability and better interpretability.

In the future, we aim to explore data-driven decision-making for crowd management, particularly concerning congested crowd dynamics, by leveraging the advantages of integrating data-driven methods with physical modeling.

\bibliographystyle{IEEEtran}
\bibliography{TCSS_reference}

\vfill

\end{document}